\documentclass[runningheads]{llncs}
\usepackage{eccv}
\usepackage{eccvabbrv}
\usepackage{graphicx}
\usepackage{booktabs}
\usepackage[accsupp]{axessibility}  
\usepackage{hyperref}

\usepackage{orcidlink}
\usepackage{multirow}
\usepackage{array} 
\usepackage{amsmath}
\usepackage{listings} %json data
\usepackage{tikz}  
\usepackage{amsfonts}
\usepackage{graphicx}
\usepackage{tcolorbox}
\usepackage{tikz}
\usepackage{caption}
\usepackage{subcaption}
\usepackage{algorithm}
\usepackage{algpseudocode}
\usepackage{booktabs}
\usepackage{tablefootnote}
\usepackage[table]{xcolor}
\usepackage{overpic}
\usepackage{tikz}
\usepackage{siunitx}
\usetikzlibrary{positioning,shapes.callouts}

\newcommand*{\belowrulesepcolor}[1]{% 
  \noalign{% 
    \kern-\belowrulesep 
    \begingroup 
      \color{#1}% 
      \hrule height\belowrulesep 
    \endgroup 
  }%
} 
\newcommand*{\aboverulesepcolor}[1]{% 
  \noalign{% 
    \begingroup 
      \color{#1}% 
      \hrule height\aboverulesep 
    \endgroup 
    \kern-\aboverulesep 
  }%
}

\newcommand*{\venue}[1]{% 
  \textcolor[rgb]{0.753,0.753,0.753}{\footnotesize \textit{#1}}
}

\begin{document}
\newcommand{\papertitle}{\textit{SATGround}}

% Variables:
\newcommand{\mx}{\mathbf{x}}
\newcommand{\mt}{\mathbf{t}}
\newcommand{\my}{\mathbf{y}}
\newcommand{\mo}{\mathbf{o}}
\newcommand{\mb}{\mathbf{b}}
\newcommand{\mz}{\mathbf{z}}
\newcommand{\mq}{\mathbf{q}}
\newcommand{\ma}{\mathbf{a}}
\newcommand{\mI}{\mathbf{I}}
\newcommand{\mell}{\boldsymbol{\ell}}
\newcommand{\mkell}{\boldsymbol{\mathit{k}}}
\newcommand{\cD}{\mathcal{D}}
\newcommand{\cM}{\mathcal{M}}
\newcommand{\cN}{\mathcal{N}}
\newcommand{\cG}{\mathcal{G}}
\newcommand{\cV}{\mathcal{V}}
\newcommand{\cT}{\mathcal{T}}
\newcommand{\cS}{\mathcal{S}}

\newcommand{\cVb}{\bar{\cV}}

\newcommand{\mepsilon}{\bm{\epsilon}}
\newcommand{\rmbin}{\mathrm{bin}}

% Variables with accents:
\newcommand{\mxh}{\hat{\mx}}
\newcommand{\mth}{\hat{\mt}}
\newcommand{\myh}{\hat{\my}}
\newcommand{\tK}{\tilde{K}}

% Variables with indexes:
\newcommand{\mxx}[1]{\mx_{#1}}
\newcommand{\mtt}[1]{\mt_{#1}}
\newcommand{\mqq}[1]{\mq_{#1}}
\newcommand{\maa}[1]{\ma_{#1}}
\newcommand{\mellell}[1]{\mell_{#1}}
\newcommand{\myy}[1]{\my_{#1}}
\newcommand{\moo}[1]{\mo_{#1}}
\newcommand{\mbb}[1]{\mb_{#1}}
\newcommand{\mll}[1]{\ml_{#1}}
\newcommand{\mxi}{\mxx{i}}
\newcommand{\mti}{\mtt{i}}
\newcommand{\myi}{\myy{i}}
\newcommand{\moi}{\moo{i}}
\newcommand{\mbi}{\mbb{i}}
\newcommand{\mqi}{\mqq{i}}
\newcommand{\mai}{\maa{i}}
\newcommand{\mli}{\mll{i}}
\newcommand{\melli}{\mellell{i}}
\newcommand{\mxhi}{\mxh_i}
\newcommand{\mthi}{\mth_i}
\newcommand{\myhi}{\myh_i}
\newcommand{\mxt}{\mxx{t}}

% Auxiliary
\newcommand{\bbR}{\mathbb{R}}
\newcommand{\bbN}{\mathbb{N}}
\newcommand{\pz}{\phantom{0}}
\newcommand{\token}[1]{\text{\textless #1\textgreater}}
\newcommand{\acc}{\text{acc}@0.5}

\def\loc{ \textcolor{red}{\<test\>} }
\newcommand{\tokenv}[1]{\langle #1 \rangle}
\title{SATGround \includegraphics[width=1.05cm, height=0.95cm]{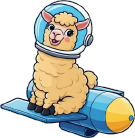} : A Spatially-Aware Approach for Visual Grounding in Remote Sensing}
\titlerunning{SATGround}
\author{Aysim Toker\inst{1} \and
Andreea-Maria Oncescu\inst{1} \and
Roy Miles\inst{1} \and
Ismail Elezi\inst{1} \and
Jiankang Deng\inst{2}}
\authorrunning{A. Toker et al.}
\institute{Huawei London Research Center \and
Imperial College London
}

\maketitle
\begin{abstract}
Vision-language models (VLMs) are emerging as powerful generalist tools for remote sensing, capable of integrating information across diverse tasks and enabling flexible, instruction-based interactions via a chat interface.
In this work, we enhance VLM-based visual grounding in satellite imagery by proposing a novel structured localization mechanism. Our approach involves finetuning a pretrained VLM on a diverse set of instruction-following tasks, while interfacing a dedicated grounding module through specialized control tokens for localization.
This method facilitates joint reasoning over both language and spatial information, significantly enhancing the model's ability to precisely localize objects in complex satellite scenes. We evaluate our framework on several remote sensing benchmarks, consistently improving the state-of-the-art, including a $33.2\%$ relative improvement over previous methods on visual grounding. Our results highlight the benefits of integrating structured spatial reasoning into VLMs, paving the way for more reliable real-world satellite data analysis. Code will be released upon acceptance.
\end{abstract}   
\section{Introduction}
\label{sec:intro}
The rapid proliferation of Earth observation technologies in recent years has provided unprecedented access to high-resolution satellite imagery, enabling advanced monitoring and analysis of the Earth’s surface. Extracting actionable insights from this vast data depends on scalable machine learning approaches. Open data initiatives like Landsat~\cite{woodcock2008free} and Copernicus~\cite{aschbacher2017esa} have facilitated the curation of specialized datasets for a wide range of applications, including disaster response, food security, and urban growth analysis. These applications require models capable of performing complex tasks, such as object recognition, temporal change detection, and interpreting scene compositionality.

Traditional machine learning approaches tackle these challenges by defining discriminative models for specific tasks such as scene classification or semantic segmentation. More recently, vision-language models (VLMs) have opened up promising new applications for multi-modal satellite data. The advantage of such generalist models lies in large-scale pretraining on image-text pairs and finetuning on instructional prompts. Hence, they can leverage prior knowledge about real-world scenes and reason about their content in a dialogue-based format, allowing them to respond to complex user queries. This conversational paradigm, where users issue queries and AI assistants generate responses, enables the joint learning of linguistic and visual semantics. A key question is how we can ensure natural interaction with these models while obtaining precise and accurate information about a given satellite scene.

\begin{figure}[tb]
    \centering
    % FIGURE 1 (Query & Legend) - Resized
    \begin{overpic}[width=0.16\textwidth]{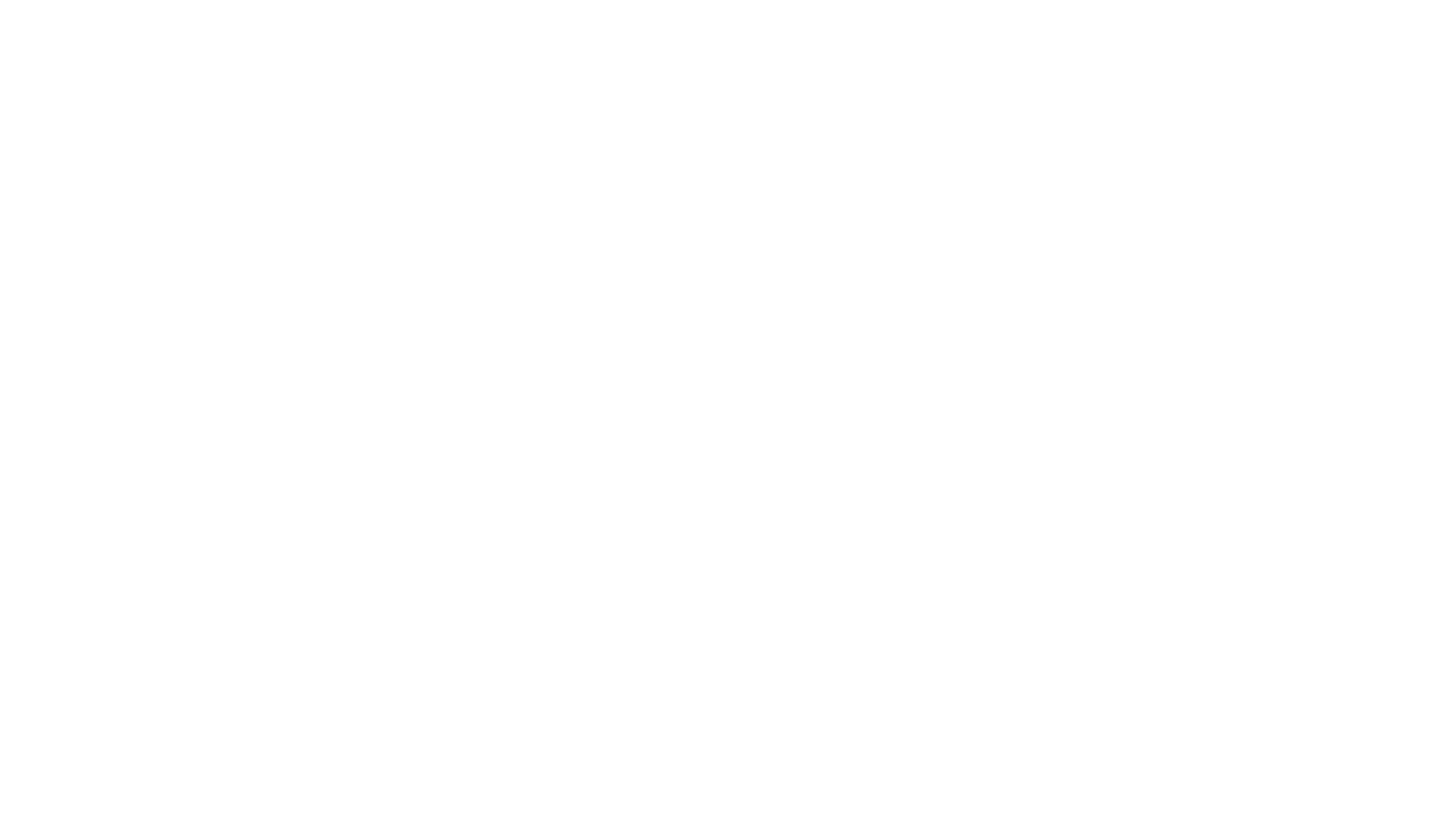} 
        \put(-15,-8.5){
        \scalebox{0.7}{
            \begin{tikzpicture}
                \node (man) {\includegraphics[height=1.4cm]{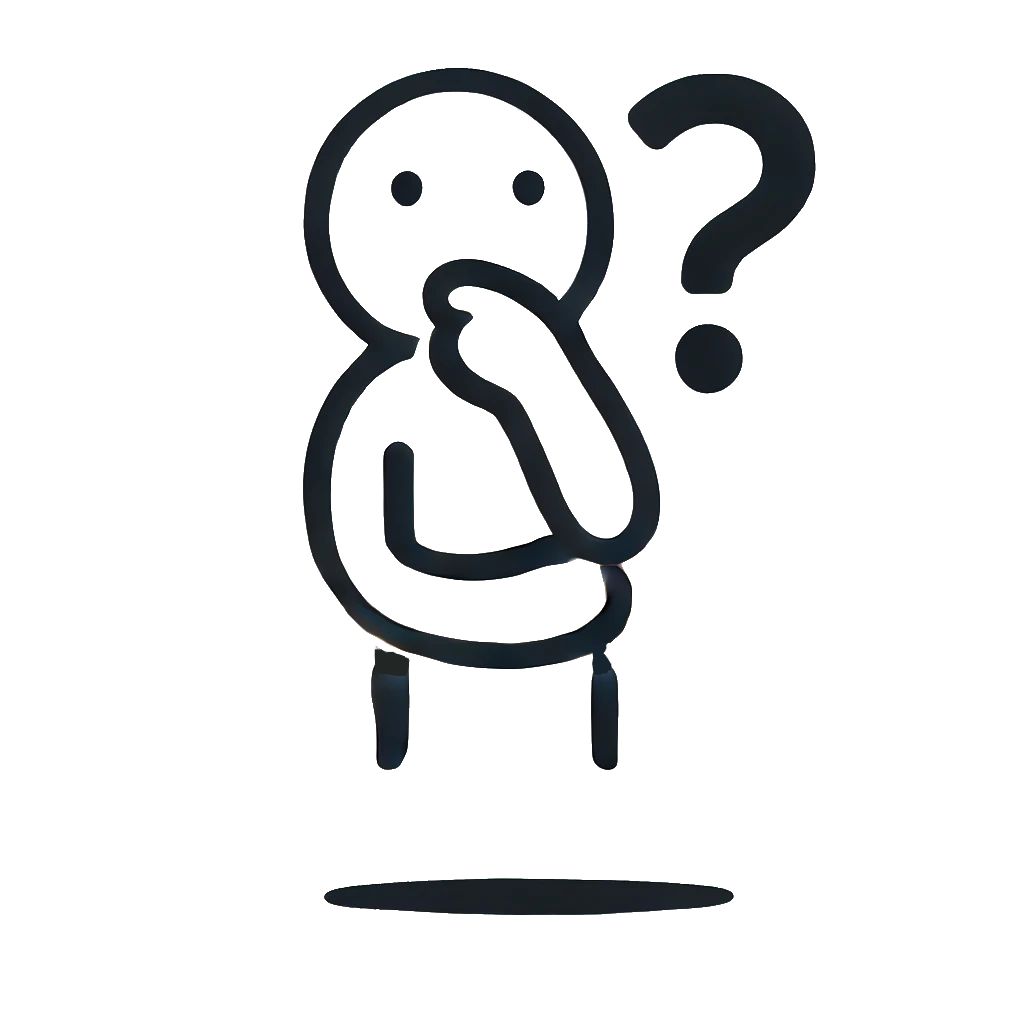}};
                \node [draw, align=center, 
                       rectangle callout, 
                       callout pointer segments = 1, anchor = pointer,
                       callout absolute pointer={(man.north east)},
                       minimum width=1.0cm, minimum height=1cm, 
                       above right = 0.1cm and -1.7cm of man.north east] 
                       {Give me the location of \\ $<p>$\dots$</p>$};
            \end{tikzpicture}}
        }
        \put(24,2){%
        \scalebox{0.55}{
        \hspace{3em}
        \begin{tikzpicture}
            \draw[green, thick] (0.5,-0.5) -- (1,-0.5) node[right] {Ours};
            \draw[orange, thick] (0.5,-0.9) -- (1,-0.9) node[right] {InternVL};
            \draw[yellow, thick] (0.5,-1.3) -- (1,-1.3) node[right] {EarthDial};
            \draw[red, thick] (0.5,-1.7) -- (1,-1.7) node[right] {g.t.};
        \end{tikzpicture}}
    }
    \end{overpic}
    \hspace{1em}
    % FIGURE 6 (NEW) - Added and resized
    % \fontsize{1}{2}\selectfont
    \begin{overpic}[width=0.145\textwidth]{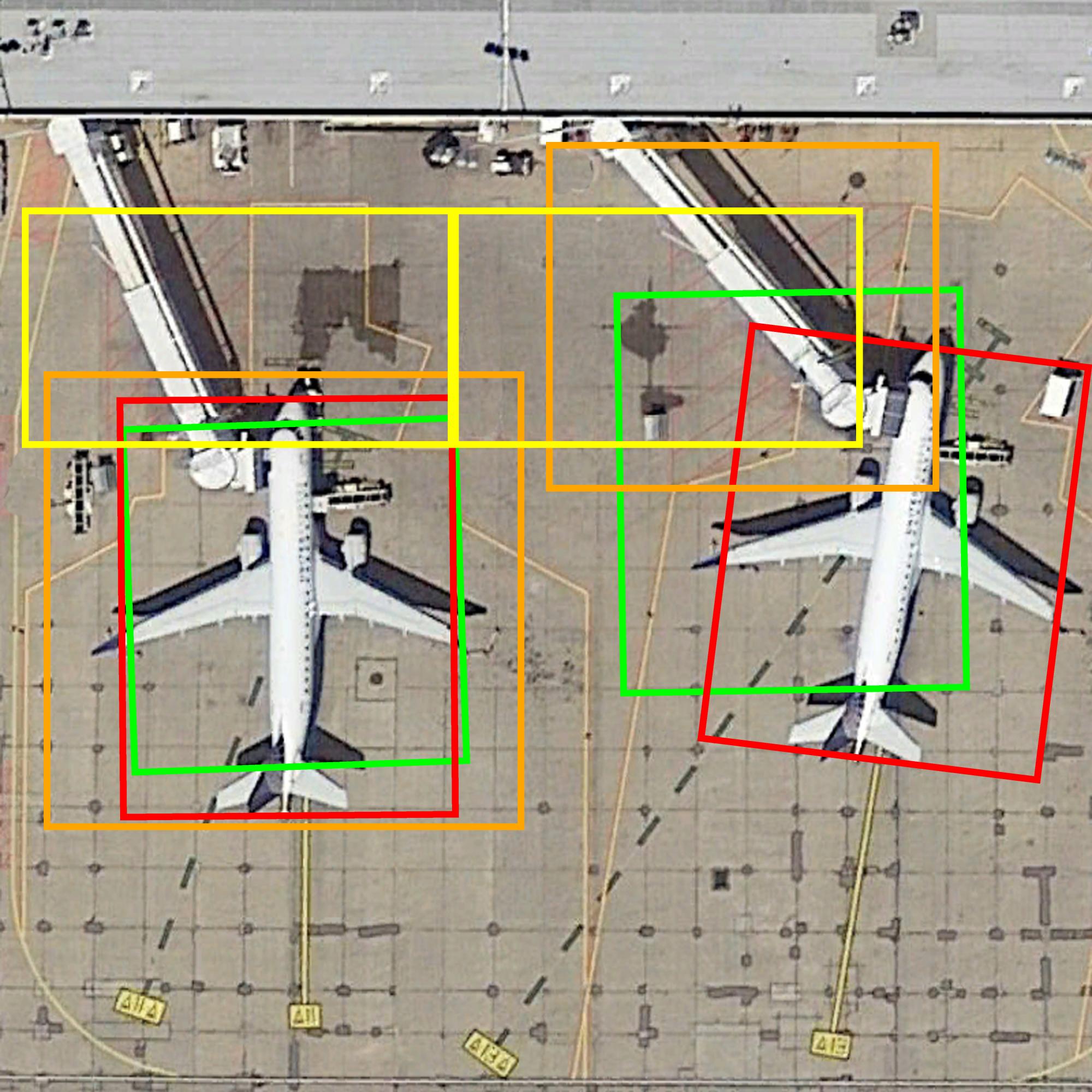}
    \put(0,-12){\begin{minipage}[t]{0.145\textwidth}\tiny\centering ``silver airplanes"\end{minipage}}
    \end{overpic}
    % FIGURE 2 - Resized
    \begin{overpic}[width=0.145\textwidth]{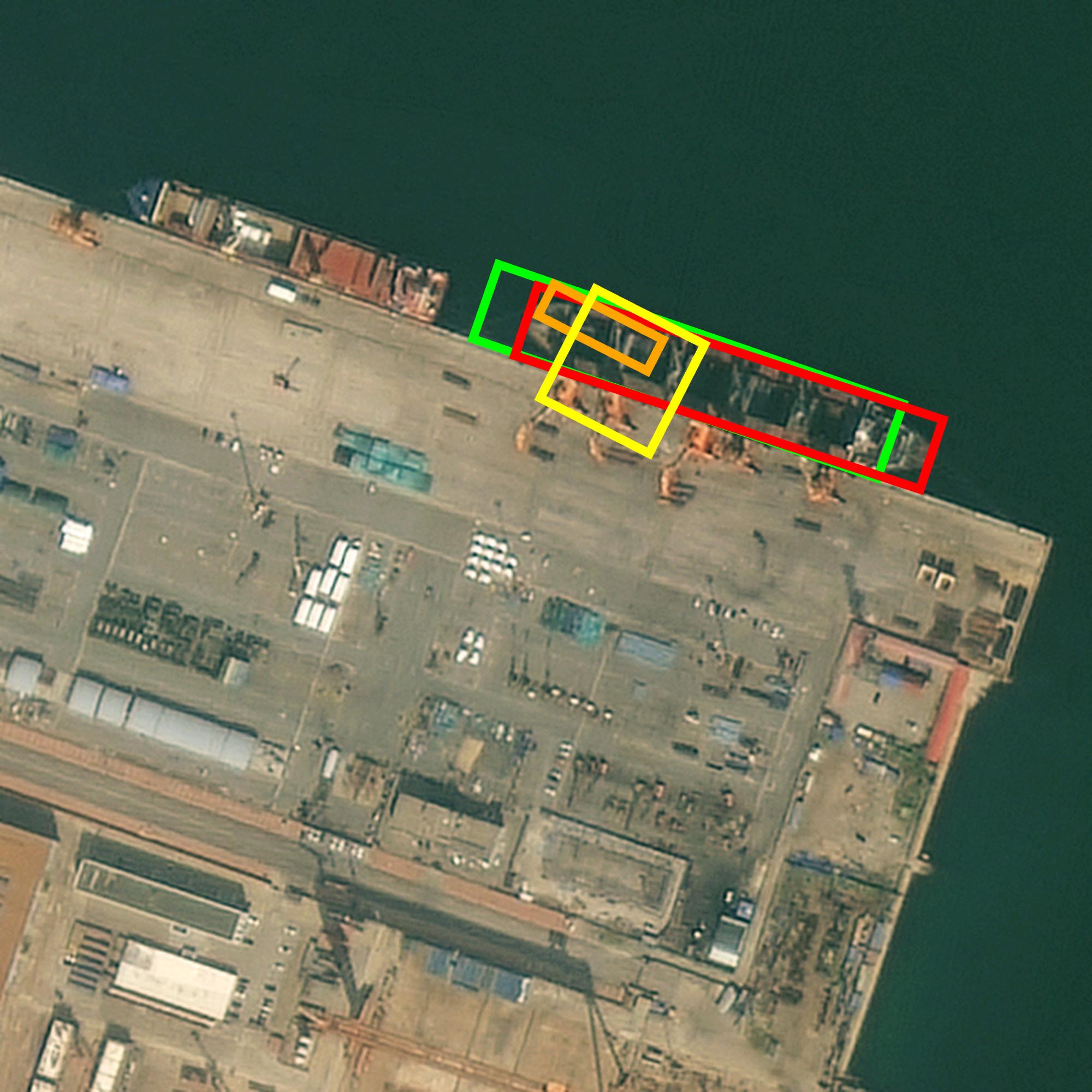}
    \put(-14,-12){\begin{minipage}[t]{0.185\textwidth}\tiny\centering ``1 dry-cargo-ship\\at the center"\end{minipage}}
    \end{overpic}
    % FIGURE 3 - Resized
    \begin{overpic}[width=0.145\textwidth]{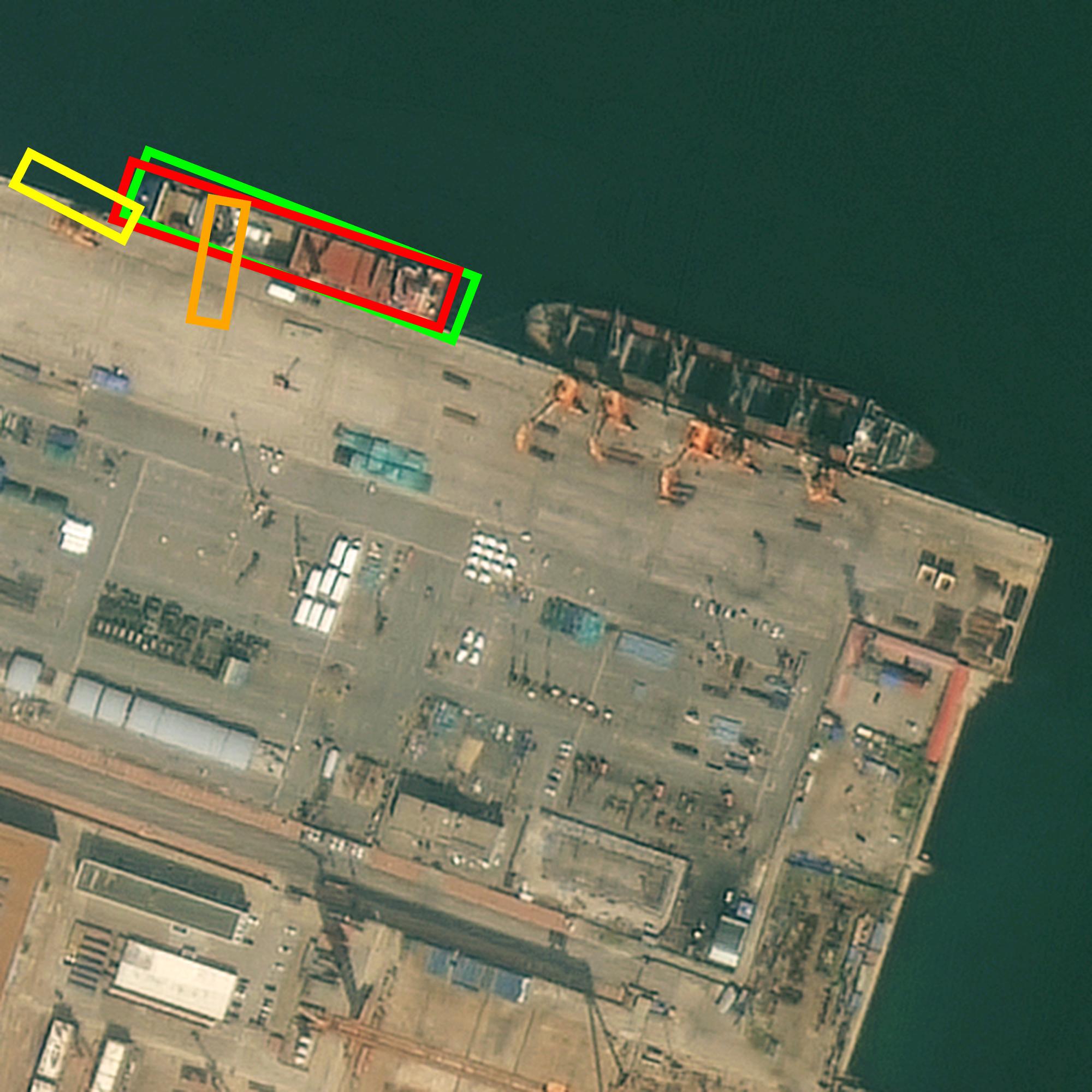}
    \put(-14,-12){\begin{minipage}[t]{0.185\textwidth}\tiny\centering ``1 dry-cargo-ship\\at the top left"\end{minipage}}
    \end{overpic}
    % FIGURE 4 - Resized
    \begin{overpic}[width=0.145\textwidth]{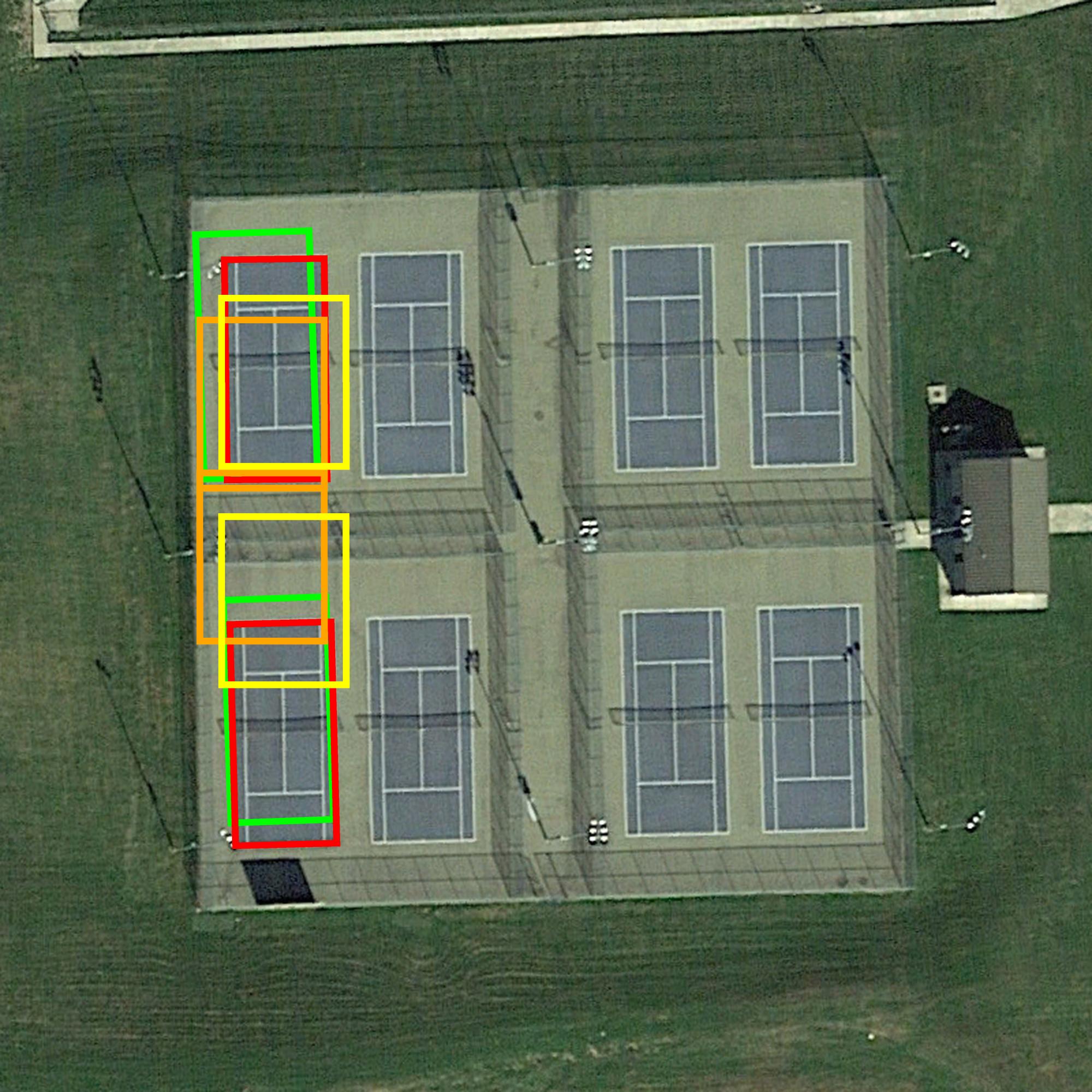}
    \put(0,-12){\begin{minipage}[t]{0.145\textwidth}\tiny\centering ``2 tenniscourts at the left"\end{minipage}}
    \end{overpic}
    % FIGURE 5 - Resized
    \begin{overpic}[width=0.145\textwidth]{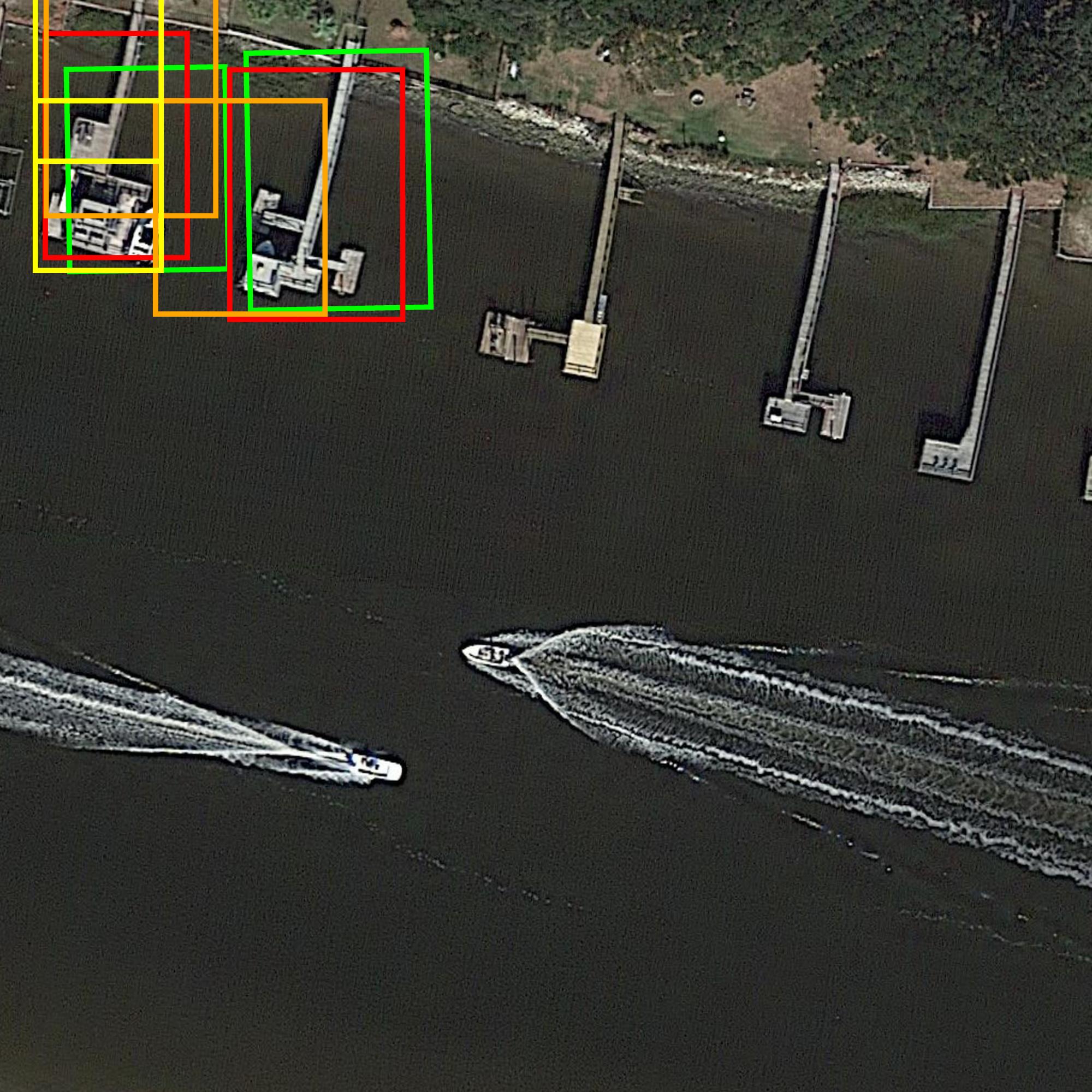}
    \put(0,-12){\begin{minipage}[t]{0.145\textwidth}\tiny\centering ``2 large harbors, at the top left"\end{minipage}}
    \end{overpic}
    
    \vspace{2.1em} 
    \caption{We propose a novel spatially-aware grounding mechanism for vision-language models (VLMs) in remote sensing. Our approach (green) integrates structured spatial information, leading to significantly improved localization capabilities compared to methods that treat bounding boxes as text. The qualitative examples shown here highlight our model's superior alignment with the ground truth (red) on diverse user prompts, outperforming state-of-the-art baselines InternVL~\cite{chen2024internvl}, CVPR'24 (orange), and EarthDial~\cite{Soni_2025_CVPR}, CVPR'25 (yellow).}
                \vspace{-1.5em} % Shrinks space before the bottom rule

    \label{fig:teaser}
\end{figure}
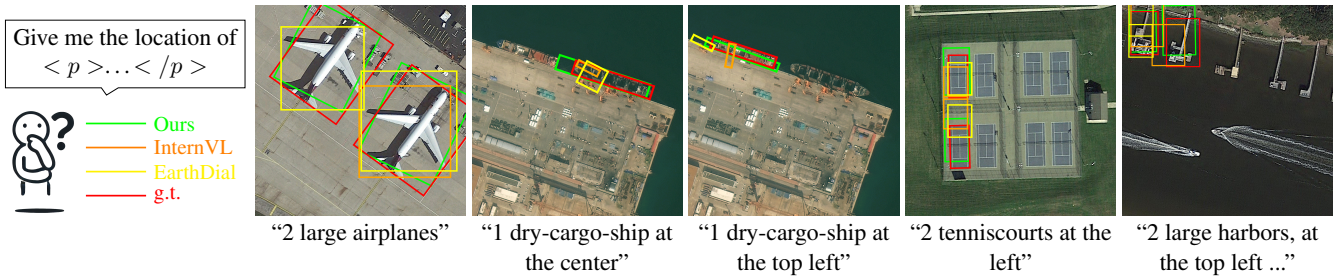

One common approach is visual grounding for detection, which reinforces the link between textual descriptions of objects and their corresponding visual representation. 
Following the pioneering approach GeoChat~\cite{cvpr2024geochat}, a number of recent works aim to leverage visual grounding for remote sensing tasks through bounding box detection~\cite{pang2024h2rsvlm,zhan2025skyeyegpt,muhtar2024lhrs,Soni_2025_CVPR}. A common strategy in these approaches is embedding bounding box locations into the prompt by converting them into text form, which may also require assigning special tokens for each possible numeric value. This is straightforward and convenient for interfacing off-the-shelf vision-language models, but it also has significant limitations due to the lack of geometric structure associated with such numerical tokens~\cite{schwartz2024numerologic,shen2023positional,singh2024tokenization}. Without an explicit spatial metric, the model does not have a notion of proximity, decreasing its robustness to small perturbations and out-of-distribution generalization. We show that this can inhibit the localization performance in practice, indicating a demand for more specialized grounding approaches.

In this work, we introduce \papertitle, a novel grounding technique for vision-language models that integrates multi-modal learning with a specialized localization mechanism for remote sensing data. We adapt existing instruction-following datasets by introducing special tokens which act as anchors for bounding box prediction and localization. Rather than converting bounding boxes into a text representation, our LLM decoder operates simultaneously in language space and bounding box space. This dual approach combines visual and textual semantics while facilitating communication between these two modalities. 

\papertitle~is a generalist framework that can be adapted to different remote sensing applications, making it both scalable and highly flexible. Our structured visual grounding technique bridges the gap between language and spatial reasoning, thus improving the localization robustness and consistency across diverse satellite imagery datasets compared to existing grounding approaches.
In summary, our \textbf{contributions} are the following:
\begin{itemize}
    \item We \textbf{propose} \papertitle, a novel, spatially-aware localization technique of visual grounding for Earth observation data.
    \item We \textbf{finetune} an existing generalist vision-language model for both language and detection objectives, interfacing these two modalities via special control tokens and a custom localization module.
    \item We \textbf{advance} the state-of-the-art on several remote sensing benchmarks~\cite{cvpr2024geochat,lobry2020rsvqa,gupta2019creating,cheng2014multi,Soni_2025_CVPR}, achieving a significant relative improvement of $33.2\%$ in visual grounding performance compared to existing methods.
\end{itemize}
\section{Related work}
\label{sec:related_work}
\noindent{\textbf{Vision-language models.} 
Over the last few years, multi-modal learning for vision tasks has become an extensively studied topic, with various approaches and applications. We provide an overview of techniques most relevant to our work here and refer to a recent survey~\cite{zhang2024vision} for a more comprehensive review. 

Early multi-modal approaches~\cite{radford2021learning,yuan2021florence,lu2019vilbert,yao2021filip,jia2021scaling,zhai2023sigmoid} revolve around a common paradigm where, given large-scale text-image pairs~\cite{schuhmann2022laion,changpinyo2021conceptual}, the model aims to integrate features into a joint embedding space with techniques such as contrastive learning. Once the embeddings are learned, this shared space is then leveraged for zero-shot learning across a variety of tasks, such as classification~\cite{radford2021learning,zhai2023sigmoid}, retrieval~\cite{plummer2015flickr30k,lin2014microsoft,Toker_2021_CVPR,DBLP:journals/pami/EleziSWVTPL23}, object detection~\cite{lin2014microsoft, DBLP:conf/cvpr/AlexandridisED025}. Later, numerous extensions were proposed that apply such models to open-vocabulary detection~\cite{gu2021open,minderer2022simple,du2022learning, fomenko2022}, segmentation~\cite{xu2022simple,Toker2022DynamicEarthNet,Toker2024SatSynthAI}, or 3D open-world learning~\cite{zhang2022pointclip,zhu2023pointclip}. These methods focus on pretraining base models and thus require finetuning for  discriminative tasks.

More recently, models like LLaVA~\cite{liu2024visual} and GPT-4~\cite{achiam2023gpt,zhu2023minigpt4} have introduced assistants capable of processing multi-modal inputs, as well as generating task-specific outputs. LLaVA~\cite{liu2024visual} provides a general framework that combines language instructions and visual embeddings into a single sequence, enabling the use of a pretrained LLM~\cite{taori2023stanford,chiang2023vicuna,peng2023instruction,touvron2023llama,goktepe2025ecomapper} as a decoder. In this work, we adopt a similar strategy, leveraging a pretrained LLM as a decoder to ground a vision-language model for Earth observation tasks.
%This approach was subsequently extended to videos~\cite{maaz2023video,li2023videochat,zhang2023video,han2023imagebind,lin2023video}

\noindent{\textbf{Grounding vision-language models.}
%\wip
The goal of visual grounding for localization is to improve the ability of models to identify the location of objects based on text-based queries.
% GLIP~\cite{li2022grounded}, GLIPv2~\cite{zhang2022dino}, DINO~\cite{zhang2022dino}, Grounding-DINO~\cite{liu2024grounding}
Training a VLM typically consists of two stages: a pretraining stage and a visual instruction tuning stage. Several works have proposed to integrate grounding into this second stage by embedding location information into a text representation~\cite{chen2023shikra,wang2023cogvlm,peng2023kosmos,you2023ferret}. This has led to many refined instruction tuning datasets~\cite{li2022grounded, zhang2022dino, chen2023shikra, peng2023kosmos} with a focus on referring and grounding tasks. 
%
% Similarly, the use of attention scores~\cite{he2024multimodalinstructiontunedllms} has been proposed to provide pixel-wise supervision.
%
However, a common limitation of all models trained in this way is that they lack any explicit geometric modeling and are contingent on the model's ability to interpret text-based location information. Prior work has shown~\cite{schwartz2024numerologic,shen2023positional,singh2024tokenization} that language models often struggle with numerical inference tasks and spatial understanding due to challenges arising from default tokenization, positional encoding schemes, and limitations in multi-step reasoning.

A number of works~\cite{zhao2023bubogpt,liu2024llava,lai2024lisa,zhang2024llava} attempted to address this problem by proposing more specialized grounding modules.
Unfortunately, these techniques are often highly engineered and specific to common object-centric data~\cite{lin2014microsoft}.
In contrast, we propose a simple new grounding technique for remote sensing data -- interfacing a generalist VLM via specialized task tokens and leveraging its rich visual semantic features to predict bounding box coordinates.

\noindent{\textbf{Vision-language models for satellite imagery.} 
Applying existing VLMs to Earth observation tasks presents a unique set of challenges, as the large domain gap between satellite and natural images often limits their performance on remote sensing data. For an overview of recent works aimed at adapting general multi-modal methods for satellite tasks, we refer to the recent survey by Li et al.~\cite{li2024vision}. Such approaches leverage task-specific datasets to enhance performance in zero-shot classification~\cite{qiu2022open,li2023rs}, image captioning~\cite{zia2022transforming}, and visual question answering~\cite{bazi2022bi,yuan2022easy,chappuis2022prompt}. However, these models are typically limited to specific tasks and lack the ability to engage in instruction-following dialogues.

The pioneering work RSGPT~\cite{hu2023rsgpt} introduces the RSICap dataset to finetune multi-modal models for instruction-following capabilities on remote sensing data, with applications such as image captioning and visual question answering (VQA). More recently, GeoChat~\cite{cvpr2024geochat} proposed to adapt the LLaVA~\cite{liu2024visual} model to remote sensing data, while integrating special task tokens into the textual instructions. This improves the zero-shot performance across a broad set of tasks, including image and region captioning, VQA, scene classification, and grounding descriptions. Following this trend, similar approaches such as SkyEyeGPT~\cite{zhan2025skyeyegpt}, LHRS-Bot~\cite{muhtar2024lhrs}, and VHM~\cite{pang2024h2rsvlm} employ comparable instruction-following architectures. The most recent efforts, such as EarthGPT~\cite{zhang2024earthgpt} and EarthDial~\cite{Soni_2025_CVPR}, introduce multi-modal models and datasets that fuse multiple data forms, such as optical, SAR, and multi-spectral.

While these works show the promise of grounded conversation in remote sensing, they often struggle to ground objects due to their reliance on encoding location context into text.
In our work, we address this limitation by introducing a dual approach that jointly models both language and bounding box spaces, enabling more precise and interpretable grounding in remote sensing data.
\section{Method}
\label{sec:method}
\subsection{Problem formulation}\label{subsec:problemstatement}
The basic framework for grounded multi-modal learning on Earth observation data can be specified as follows:
\begin{equation}\label{eq:generalvlm}  
    \cM_\text{vlm}: (\mx, \mq) \mapsto \ma,
\end{equation}  
where the vision-language model $\cM_\text{vlm}$ processes a satellite image $\mx \in \bbR^{H \times W \times 3}$ and a text query $\mq \in \cV^{N_q}$ to produce a textual response $\ma \in \cV^{N_a}$. We commonly assume that addressing $\mq$ requires reasoning about the content of the image $\mx$. Here, $W,H\in\bbN$ denote the spatial size of the input image, while $N_q,N_a\in\bbN$ correspond to the sequence length of $\mq$ and $\ma$, respectively. Both sequences specify tokens from a common vocabulary:
\begin{equation}\label{eq:vocabulary}  
   \cV:=\cT\cup\cS:=\{\mt_1,\dots,\mt_{|\cT|}\}\cup\{\tokenv{bos}, \tokenv{eos}, \tokenv{pad}\},
\end{equation}
which comprises text tokens $\cT$, as well as special control tokens $\cS$, such as the beginning-of-sequence token $\langle bos \rangle$, the end-of-sequence token $\langle eos \rangle$, and the padding token $\langle pad \rangle$. 

\noindent{\textbf{Motivation.}} Standard grounding techniques for vision-language models on Earth observation data embed location context within text representations, to allow implicit learning of spatial relations through next-token prediction. However, this method lacks explicit geometric modeling, which hinders the ability of the model to capture the precise spatial layout of satellite images.

To address this limitation, we modify the framework to incorporate explicit geometric reasoning. Rather than relying on implicit location information via next-token prediction, we introduce a grounding mechanism to predict bounding box locations directly. This model is defined as follows:
\begin{equation}\label{eq:ourvlm}
    \cM_\text{loc}: (\mx, \mq) \mapsto (\ma, \mell).
\end{equation}

\begin{figure}[t!]
    \includegraphics[width=1.\textwidth, keepaspectratio]{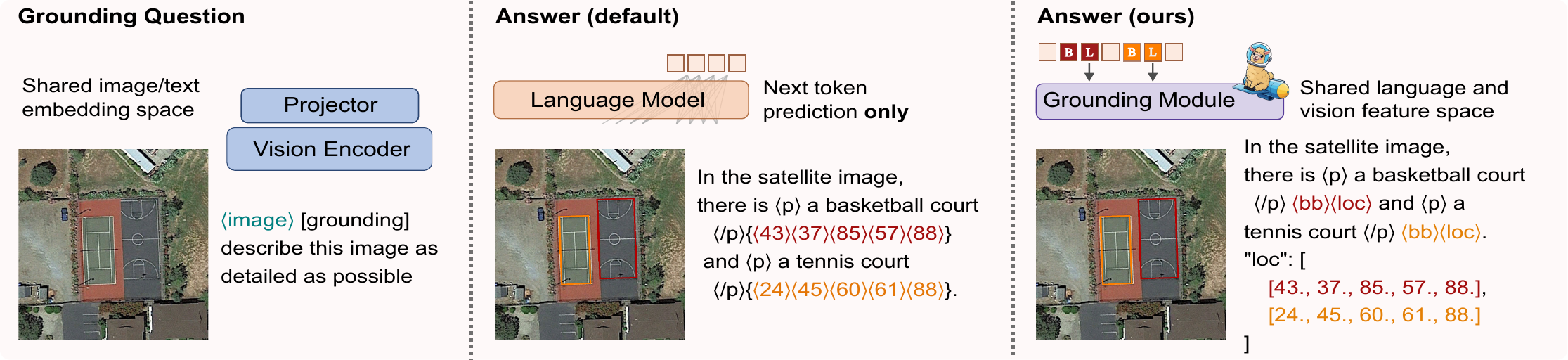}
    \caption{\textbf{Structured visual grounding.} For a given user query (left), we visualize the conventional text-based visual grounding approach (middle), compared to our structured explicit grounding format (right). For illustration purposes, we show the ground-truth bounding box location values (red and orange) in both cases. Instead of returning bounding box coordinates as text, we model a dedicated localization mechanism interfaced by special control tokens $\tokenv{bb}$ and $\tokenv{loc}$, see~\cref{subsec:linkingregression} for more details. 
    }
                \vspace{-1.5em} % Shrinks space before the bottom rule
\label{fig:ourresponse} 
\end{figure}

Besides mapping an image $\mx \in \mathbb{R}^{H \times W \times 3}$ and text query $\mq \in \cVb^{N_q}$ to a reply $\ma \in \cVb^{N_a}$, this localization model $\cM_\text{loc}$ also directly predicts the bounding box locations $\mell \in \mathbb{R}^{L \times 5}$.
Here, $\mell$ represents coordinates of $L\in\bbN$ different predicted bounding boxes, see~\cref{sec:implementationdetails} for bounding box specifications. We further append two additional special tokens to the vocabulary:
% each fully specified by $5$ parameters: $4$ for the location (coordinates of the top left corner, bottom right corner) and $1$ for the orientation (angle of rotation). 
\begin{equation}\label{eq:vocabulary_additional}  
   \cVb:=\cV\cup\{\tokenv{bb},\tokenv{loc}\},
\end{equation}
denoting bounding boxes $\tokenv{bb}$ and location $\tokenv{loc}$ tokens, respectively. In~\cref{fig:ourresponse}, we visualize our bounding box representation for a given query in comparison to the standard text-based format. We describe in~\cref{subsec:linkingregression} how integrating these tokens enables structured visual grounding in our model.

To facilitate gradient-based learning, we define $\cM_\text{loc}$ with a differentiable mapping from image $\hat{\mx}$ and text $\hat{\mq}$ embeddings to locations $\mell$. By convention, we denote $\hat{\mx}$ and $\hat{\mq}$ for the embedded features in latent space corresponding to data instances such as images $\mx$ and text $\mq$. The model thus optimizes the language output and bounding box locations simultaneously, enabling it to reason about how geometric details correspond to textual cues.

\subsection{Structured visual grounding}\label{subsec:linkingregression}
The vocabulary space of possible output tokens in~\cref{eq:vocabulary_additional} includes text tokens, control tokens $\langle bos \rangle, \langle eos \rangle, \langle pad \rangle$, and two additional special tokens $\langle bb \rangle$ and $\langle loc \rangle$. We now detail the latter two, which specify a grounding interface for regressing bounding box parameters in our model. Instead of forcing the language model to output numerical coordinates as raw text, this interface routes spatial predictions to a dedicated grounding module.

\begin{figure}[t!]
    \centering
    \includegraphics[width=\linewidth]{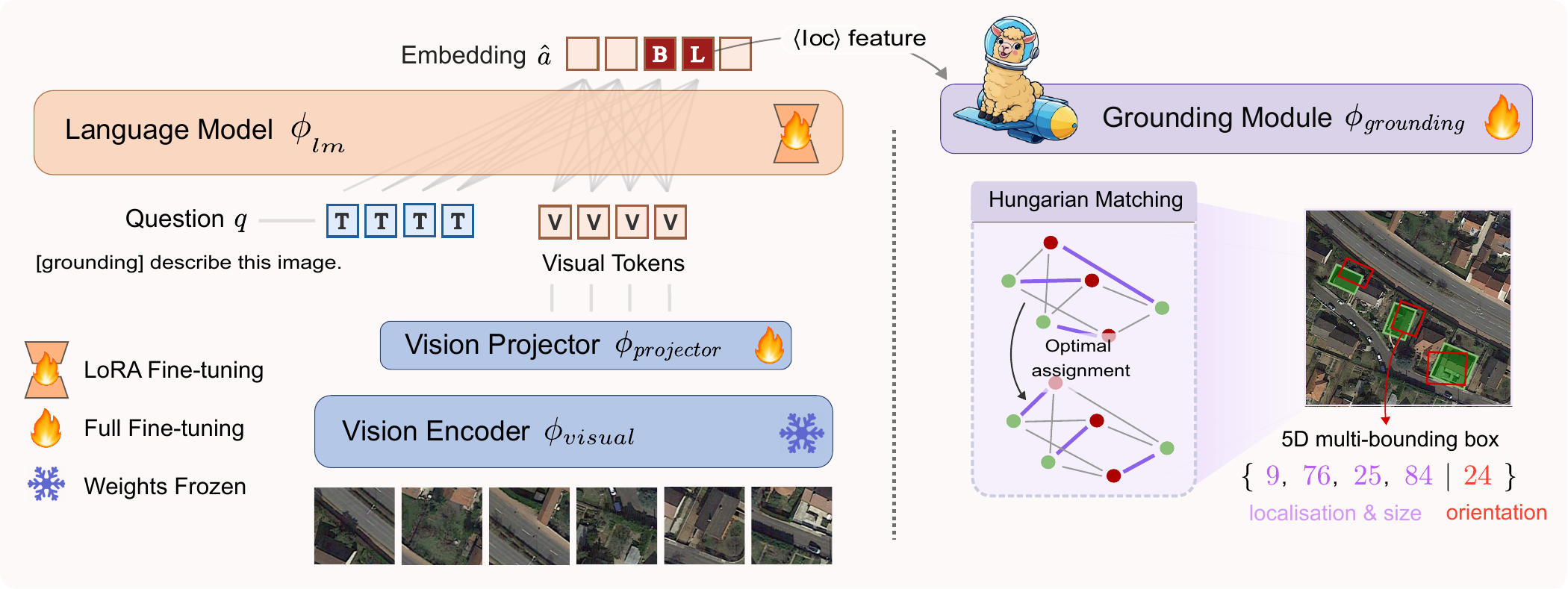}
    \caption{\textbf{Method overview.} We summarize the different components of our model. Visual inputs are encoded using a frozen vision backbone, $\phi_\text{visual}$, and a trainable adapter $\phi_\text{projector}$. The extracted vision tokens are then concatenated with the user query and passed to the language model $\phi_\text{lm}$ based on LLaVA~\cite{liu2024visual}, which we finetune by applying LoRA~\cite{hu2022lora}. The model's response contains both standard text tokens and occasional $\tokenv{bb}$ and $\tokenv{loc}$ tokens, see~\cref{subsec:linkingregression} for more details. Any resulting $\tokenv{loc}$ feature embeddings are passed to the grounding module $\phi_\text{grounding}$ which produces the final bounding box predictions and matches them by using the Hungarian algorithm.}
    \vspace{-1.5em} % Shrinks space before the bottom rule
    \label{fig:overview_fig}
\end{figure}

\noindent{\textbf{Bounding box and location tokens.}}
By convention, the two special bounding box tokens always appear in sequence $\langle bb \rangle\langle loc \rangle$ in a given text sample. They serve two distinct purposes: $\langle bb \rangle$ indicates that the model produces a bounding box as part of the current predicted sentence. $\langle loc \rangle$ then provides latent features for the corresponding bounding box parameters (\ie, its location and orientation). The design choice of introducing two separate tokens $\langle bb \rangle$ and $\langle loc \rangle$ ensures a clear separation of responsibilities, and avoids sharing the capacity of a single embedding vector between these tasks. Our ablation study in \cref{tab:ablationstudy}(a) confirms this design is crucial, demonstrating that the separated tokens achieve an improved performance compared to a single-token approach.

\noindent{\textbf{Training.}} During training, our model naturally produces bounding box tokens $\ma_t=\langle bb \rangle$ as part of the output sequence $\ma\in\cVb^{N_a}$ via standard next-token prediction. In contrast, the location tokens $\langle loc \rangle$ are masked out when computing the cross-entropy token loss, analogous to padding tokens $\langle pad \rangle$. Since $\langle loc \rangle$ always appears right after $\langle bb \rangle$ in a fully deterministic manner, it carries no information for next-token prediction (zero entropy). Instead, we simply pass the latent embeddings $\hat{\ma}_t\in\mathbb{R}^{D}$ associated with the $\ma_t=\langle loc \rangle$ tokens to our grounding module $\phi_\text{grounding}$, where $D$ denotes the embedding dimensionality. This separate head is responsible for regressing the final bounding box coordinates:
\begin{equation}
\phi_\text{grounding}:\begin{cases}
\mathbb{R}^{L \times D} \to \mathbb{R}^{L \times 5} \\
\{\hat{\ma}_t\in\mathbb{R}^{D}|\ma_t=\langle loc \rangle\} \mapsto \mell
\end{cases}
\end{equation}
% which predicts the bounding box coordinates $\mell \in \mathbb{R}^{L \times 5}$. In practice $\phi_\text{grounding}$ is defined as a shallow MLP. 
where $\mell \in \mathbb{R}^{L \times 5}$ denotes the predicted bounding box coordinates. In practice, $\phi_\text{grounding}$ is parameterized as a shallow MLP.
% We follow the convention of representing each bounding box as a 5-dimensional parameter vector, indicating the coordinates of the top left corner, the bottom right corner, and its rotation. 
% In~\cref{fig:ourresponse}, we visualize our bounding box representation for a given query in comparison to the standard text-based format. 
The primary goal is to accurately predict the location of a specific object while interfacing the VLM's language capabilities. We provide a comprehensive graphical representation of our approach in~\cref{fig:overview_fig}. To finetune the model, we optimize for both the language modeling and location tasks simultaneously:
\begin{equation}\label{eq:loss}
 \mathcal{L} = \lambda_\text{text}\mathcal{L}_\text{CE} + \lambda_\text{bb}\mathcal{L}_\text{ground}, 
\end{equation}
where $\mathcal{L}_\text{CE}$ represents the cross-entropy loss for predicting the next token in the sequence, $\mathcal{L}_\text{ground}$ is the grounding loss for regressing the bounding box coordinates, and $\lambda_\text{text}, \lambda_\text{bb}$ are hyperparameters indicating their relative weights. 

\noindent{\textbf{Bipartite matching.}} Since the model may output bounding box predictions for multiple distinct visual categories in a single response, a standard global, class-agnostic bipartite matching can lead to semantically flawed assignments. For instance, in the scenario depicted in Fig. 2, the model might match a ``basketball court'' prediction to a ``tennis court'' annotation solely due to spatial proximity. To enforce permutation invariance while preserving semantic associations, we partition the predicted bounding boxes into $G$ distinct referring expression groups based on their sequential adjacency within the generated text, specifically consecutive $\langle bb \rangle$ tokens uninterrupted by descriptive text. Within each group, we compute a bipartite matching of the predicted bounding boxes $\mell\in\mathbb{R}^{L \times 5}$ to the reference bounding box coordinates $\mkell\in\mathbb{R}^{K \times 5}$ to calculate the grounding loss $\mathcal{L}_\text{ground}$. Following standard practice in the literature~\cite{carion2020end,stewart2016end}, this is formulated as a linear sum assignment:
% \begin{equation}\label{eq:hungarianloss}
%     \mathcal{L}_\text{ground}(\mell,\mkell):=\min_{\rho\in\mathcal{P}_{L,K}}\sum_{i=1}^L\sum_{j=1}^{K}\rho_{i,j}\|\mell_i - \mkell_j\|_1,
% \end{equation}
\begin{equation}\label{eq:hungarianloss}
    \mathcal{L}_\text{ground}(\mell,\mkell):=\sum_{g=1}^G\min_{\rho^{(g)}\in\mathcal{P}_{L_g,K_g}}
    \sum_{i=1}^{L_g}\sum_{j=1}^{K_g}\rho^{(g)}_{i, j} \|\mell^{(g)}_i - \mkell^{(g)}_j\|_1
    % \in\mathcal{P}_{L,K}}\sum_{i=1}^L\sum_{j=1}^{K}\rho_{i,j}\|\mell_i - \mkell_j\|_1,
\end{equation}
which can be efficiently solved using the Hungarian algorithm~\cite{kuhn1955hungarian}. Here, $\rho^{(g)}\in\mathcal{P}_{L_g,K_g}\subset\{0,1\}^{L_g \times K_g}$ denotes a partial permutation matrix that defines an optimal one-to-one matching between a subset of the predicted and reference boxes within group $g$, allowing for unmatched predictions or ground-truth boxes when their counts differ. By resolving assignment ambiguities strictly among bounding boxes referring to the same visual entity, this local formulation effectively eliminates cross-expression contamination. As demonstrated in our ablation studies~(\cref{tab:ablationstudy}(a)), enforcing this grouped, permutation-invariant matching is crucial for preventing semantic mismatches and maintaining precise localization. By isolating this spatial reasoning within a dedicated module, the primary text generation of the VLM remains unhindered and highly efficient.

\noindent{\textbf{Inference.}} During inference, we apply standard next-token prediction in a greedy sampling scheme to obtain an answer sequence $\ma\in \cV^{N_a}$ and bounding box coordinates $\mell\in\mathbb{R}^{L \times 5}$ from the model for a given query sequence $\mq \in \cV^{N_q}$ and reference image $\mx \in \bbR^{H \times W \times 3}$. Whenever the model produces a $\tokenv{bb}$ token, we do not query the LLM and instead append a $\tokenv{loc}$ token immediately afterwards in a fully deterministic manner. We further register any such occurrences of $\tokenv{loc}$ tokens and use the resulting features to query our grounding module $\phi_\text{grounding}$ at the very end.~\cref{alg:greedy_sampling} details this sampling procedure.
\begingroup
% 1. Cut the gap between your document text and the top/bottom of the algorithm box
\setlength{\intextsep}{4pt}      % Adjust from default ~12pt down to 4pt
\setlength{\textfloatsep}{4pt}   % For floats at top/bottom of page

\begin{algorithm}[h!]
    \centering
    % 2. Cut the gap between the caption and the top line of the algorithmic block
    \setlength{\belowcaptionskip}{-4pt} 
    
    \caption{\textbf{Greedy sampling.} Pseudocode for greedy sampling in our model. The model's LM head $\phi_\text{lm-head}$ selects the most probable token at each step without randomness. After all answer tokens are generated, we additionally query the grounding module $\phi_\text{grounding}$ to predict a set of bounding box parameters for each instance of a predicted location token $\ma_{t}=\tokenv{loc}$.}
    \label{alg:greedy_sampling}

    \begin{algorithmic}[1]
        % 3. To shrink space INSIDE the algorithmic block, place \vspace AFTER \begin{algorithmic}
        % \vspace{-0.2em} 
        \Statex \textbf{Input:} Image $\mx \in \bbR^{H \times W \times 3}$, query $\mq \in \cV^{N_q}$
        \Statex \textbf{Returns:} Answer tokens $(\ma_{1}, ..., \ma_{N_a}) \in \cV^{N_a}$,
        \Statex bounding box parameters $\mell\in\mathbb{R}^{L \times 5}$
        \State // \textit{Compute visual $\hat{\mx}$ and textual $\hat{\mq}$ embeddings:}
        \Statex $\hat{\mx}\leftarrow\phi_\text{projector}\circ\phi_\text{visual}(\mx)\in\bbR^{P \times D}$;\quad  $\hat{\mq}\leftarrow\phi_\text{embed}(\mq)\in\bbR^{N_q \times D}$;\quad 
         $\ma_0\leftarrow\emptyset$
        \For{$t = 1$ to $N_a$} \hfill// \textit{Iteratively generate tokens}\hspace*{40pt} {}
            \If{$\ma_{t-1} = \tokenv{eos}$}
                \State \textbf{Break} \hfill// \textit{Stop generation after EOS}\hspace*{40pt} {}
            \ElsIf{$\ma_{t-1} = \tokenv{bb}$}                \Statex\hspace{30pt}$\ma_t\leftarrow\tokenv{loc}$  \hfill//\textit{Deterministic $\tokenv{loc}$ after $\tokenv{bb}$ token}\hspace*{40pt} {}
            \Else\Statex\hspace{30pt}$\hat{\ma}_t\leftarrow\phi_\text{lm}(\hat{\mx}, \hat{\mq}|\ma_{:t})$ \hfill// \textit{Query next token from the LM}\hspace*{40pt} {}
            %\State Compute logits: $z_t = f_\theta(\hat{x}_{1:t-1})$
            \Statex\hspace{30pt}$\ma_t\leftarrow\phi_\text{lm-head}(\hat{\ma}_t)\in\cVb$ \hfill// \textit{Apply LM head}\hspace*{40pt} {}
            \EndIf
        \EndFor
        \Statex $\hat{\ma}_{\text{loc}}\leftarrow\{\hat{\ma}_t\in\mathbb{R}^{D}|\ma_{t}=\tokenv{loc}\}\in\mathbb{R}^{L \times D}$ \hfill// \textit{Extract location tokens}\hspace*{40pt} {}
        \Statex $\mell\leftarrow\phi_\text{grounding}(\hat{\ma}_{\text{loc}})$ \hfill// \textit{Query grounding module}\hspace*{40pt} {}
        \State \Return $\ma,\mell$
        % \vspace{-0.2em} % Shrinks space before the bottom rule
    \end{algorithmic}
    \vspace{-0.2em}
\end{algorithm}
\endgroup
\subsection{Implementation details}\label{sec:implementationdetails} 
% To finetune our model for temporal remote sensing data, we use the TEOCHATLAS~\cite{irvin2024teochat} dataset, which is a superset of GeoChat that incorporates samples from 4 additional earth observation benchmarks: fMoW~\cite{christie2018functional}, xBD~\cite{gupta2019creating}, S2Looking~\cite{shen2021s2looking}, and QFabric~\cite{verma2021qfabric}. These datasets cover seven distinct temporal instruction-following tasks: temporal scene classification, change detection, spatial change referring expression, change question answering, region-based change question answering, temporal referring expression, and region-based temporal question answering.
\noindent{\textbf{Oriented bounding box representation.}}
Unlike natural images, objects in satellite imagery can appear at arbitrary orientations due to the overhead viewing angle, making axis-aligned bounding boxes a poor fit. We therefore model oriented bounding boxes (OBBs) to capture both the spatial extent and the rotation of each object. Concretely, each bounding box is parameterized as a 5-tuple $(x_1, y_1, x_2, y_2, \theta)$, where $(x_1, y_1)$ and $(x_2, y_2)$ denote the top-left and bottom-right corners, and $\theta \in [0^\circ, 180^\circ)$ specifies the counter-clockwise rotation from the horizontal axis.
Following the annotation convention of GeoChat~\cite{cvpr2024geochat} and EarthDial~\cite{Soni_2025_CVPR}, coordinates and angles in the dataset are normalized to $[0, 100]$ via explicit linear mapping: $x_{norm} = 100 \cdot x / W$, $y_{norm} = 100 \cdot y / H$, and $\theta_{norm} = 100 \cdot \theta / 180^\circ$. For training, we further rescale these values to continuous targets in $[0, 1]$. To ensure the output predictions from $\phi_\text{grounding}$ strictly adhere to this valid range, we apply a sigmoid activation function to the final layer. We ablate this design choice against unconstrained linear outputs in~\cref{tab:ablationstudy}(d); while both configurations perform reasonably, the sigmoid-constrained head yields consistently higher accuracy across all metrics.
% \noindent{\textbf{Datasets.}}
% Our main motivation is to finetune a generalist VLM that can be applied to different downstream visual perception tasks on remote sensing data.
% To this end, we utilize the GeoChat~\cite{cvpr2024geochat} and EarthDial~\cite{Soni_2025_CVPR} multi-modal datasets for finetuning our model. These datasets are aggregated from a variety of sources~\cite{rahnemoonfar2021floodnet,cheng2017remote,lobry2020rsvqa} for a range of tasks, including visual question answering (VQA), grounding description, detailed description, multi-turn conversation, region captioning, referring expression, and scene classification.
% %While not all of these tasks are amenable to benchmarking (some only allow for subjective qualitative assessment)
% Training on such a large suite of problems helps to instill a holistic understanding of relevant data instances into the model.

\noindent{\textbf{Architecture details.}}
Our model's core architecture consists of a pretrained, frozen ViT visual encoder~\cite{radford2021learning}, a trainable vision projector, and a language decoder. We also employ a grounding module ($\phi_\text{grounding}$) consisting of a 2-layer MLP, which is responsible for mapping our location embeddings ($\hat{\ma}_t\in\mathbb{R}^{D}$ associated with $\ma_t=\langle loc \rangle$) to bounding box coordinates, as detailed in~\cref{subsec:linkingregression}.
To ensure a fair comparison with the GeoChat and EarthDial baseline approaches, our experiments use the same respective language decoder. Specifically, we use the Vicuna-v1.5~\cite{chiang2023vicuna} model for the GeoChat comparison and the Phi-3-mini~\cite{phi3_mini} model for EarthDial. For data processing on GeoChat, we use 504$\times$504 image inputs with interpolated positional embeddings. For the EarthDial benchmark, we employ the multi-modal (i.e., multi-spectral, SAR, temporal) and multi-resolution data fusion strategy detailed in~\cite{Soni_2025_CVPR}. Unless stated otherwise, all following experiments use this EarthDial setting. 

\noindent{\textbf{Training details.}} To develop a generalist model capable of diverse remote sensing tasks, we finetune on the GeoChat \cite{cvpr2024geochat} and EarthDial \cite{Soni_2025_CVPR} datasets. These provide a diverse suite of tasks, including VQA, grounding, region captioning, and scene classification, aggregated from multiple sources \cite{rahnemoonfar2021floodnet, cheng2017remote, lobry2020rsvqa}. We finetune our model for $1$ epoch with a batch size of $16$, using the AdamW optimizer with a learning rate of $2 \times 10^{-4}$ and a cosine annealing schedule. To keep the finetuning lightweight, we apply LoRA~\cite{hu2022lora} with rank $r=64$ and scaling factor $\alpha=16$ on the language backbone, while keeping the vision encoder frozen. The grounding module $\phi_\text{grounding}$ and the vision-language projector are trained with half precision. Our total trainable overhead amounts to only $16.8$M parameters.

\section{Experiments}
\label{sec:experiments}
% \begin{figure*}[!t]
%     \centering
%     % \vspace{1.5em}
%     \includegraphics[width=\linewidth]{figures/qualitatives_no_zoom_no_man.pdf}
%     \caption{\textbf{Qualitative comparison for visual grounding}. These instances correspond to the quantitative results from~\cref{tab:groundingperformance}. For each sample, we provide predictions by our approach in green, InternVL~\cite{chen2024internvl} in orange, and EarthDial~\cite{Soni_2025_CVPR} in yellow. The ground truth bounding boxes are shown in red.}
%                     \vspace{-1.5em} % Shrinks space before the bottom rule

%     \label{fig:qualitativegrounding}
% \end{figure*}
We evaluate our method on several tasks, including referring expression detection, grounding description, and two visual question answering benchmarks. For several of these tasks, we also assess generalization to different datasets in a zero-shot manner.
\subsection{Grounding and referring expression detection}\label{subsec:visualgrounding}
\noindent{\textbf{Dataset.}} We evaluate our model on the GeoChat~\cite{cvpr2024geochat} test set, and further assess zero-shot performance on NWPU VHR-10~\cite{cheng2014multi}, Swimming Pool~\cite{Soni_2025_CVPR} and Urban Tree Crown Detection~\cite{zamboni2021benchmarking}. These four datasets comprise 7,593, 1,601, 5,697, and 1,279 instances, respectively, all of which feature a mix of single and multi-object queries. The benchmarks differ in their task definition: by design, GeoChat categorizes queries into distinct ``Grounding” and ``Referring” groups based on object attributes, while the remaining three benchmarks feature a single, holistic ``Referring” task. This diverse selection of datasets ensures our model is rigorously tested across a wide variety of object distributions and scales.

\noindent{\textbf{Task.}} The objective here is to precisely localize the visual objects specified by a text query $\mq$ within an input image $\mx$, outputting their corresponding bounding box locations $\mell$. The queries in the considered benchmarks refer to diverse objects of interest, including storage tanks, cars, and airplanes, swimming pools, as well as larger objects such as football fields and bridges. We follow the prior convention~\cite{cvpr2024geochat,Soni_2025_CVPR} of reporting the $\acc$ metric as the main evaluation score, defined as the fraction of correct predictions with an IoU score larger than $0.5$.
% \begin{table*}[!t]
% \vspace{1.1em}
% \begin{center}
% \scalebox{0.90}{
% % \footnotesize
% \begin{tabular}{lcccccccc}
% \toprule
%  & Small & Medium & Large & Single [g] & Multi [g] & [refer] & [grounding] & Overall \\
% \toprule
% LHRS-Bot~\cite{muhtar2024lhrs} &  &  &  &  &  &  &  &  \\
% GeoChat~\cite{cvpr2024geochat}&  7.9& 29.5 & 36.4& 39.0 & 19.3 &  21.8 & 20.6 &  21.7\\
% TeoChat~\cite{irvin2024teochat} &  &  &  &  &  &  &  &  \\
% MiniGPTv2~\cite{chen2023minigptv2}& 1.7 & \pz9.9 & 21.9 & \pz9.1 & 3.6 & \pz8.2 & 2.6 & \pz7.6  \\
% Ferret~\cite{you2023ferret} & 7.4 & 24.5 & 31.5 & 23.7 & 9.1 & 19.3 & 10.4 & 18.7 \\
% InternVL2-4B~\cite{chen2024internvl}&  \pz7.1 & 23.3 & 35.2 &  35.6 & 11.7 & 19.7 &12.9  & 19.0 \\
% EarthDial~\cite{Soni_2025_CVPR}&  \pz7.0 & 24.7 &  34.6&  33.9&  12.7&  20.1& 13.8 & 19.4 \\
% % \hline
% \rowcolor{cyan!15} Ours & \textbf{10.3} & \textbf{34.2} &  \textbf{52.6} &  \textbf{40.7} &  \textbf{19.4} &  \textbf{28.8} &  \textbf{20.8} &  \textbf{28.1} \\
% \aboverulesepcolor{cyan!15}
% \midrule[0.1em]
% \end{tabular}
% }
% \centering
% \caption{\textbf{Quantitative results of visual grounding.\textcolor{purple}{precision(OPTIMAL EVAL MODE PRECISION )!!!}} }
% \label{tab:groundingprecisionperformance_optimal}
% \end{center}
% \vspace{-0.5cm}
% \end{table*}

\begin{table}[!t]
    \centering
     \small
    \caption{
        \textbf{Quantitative results for visual grounding and referring on the GeoChat benchmark.} We report the accuracy ($\%$) score. Performance is broken down by object size (Small, Medium, Large) and task type. For the Grounding and Referring tasks, we evaluate performance on prompts with single vs. multiple objects. The Avg. column shows the mean score for each task category. 
    }
    \vspace*{-.4\baselineskip}
    \label{tab:groundingperformance}
\resizebox{0.85\textwidth}{!}{
     \begin{tabular}{lccccccccccc}
        \toprule
        & \multicolumn{3}{c}{Object Size} & \multicolumn{3}{c}{Grounding} & \multicolumn{3}{c}{Referring} & {\multirow{2}{*}{Overall}} \\ 
        \cmidrule(lr){2-4} \cmidrule(lr){5-7} \cmidrule(lr){8-10}
        Model & {Small} & {Medium} & {Large} & {Single} & {Multi} & {Avg.} & {Single} & {Multi} & {Avg.} & \\
        \midrule
        \venue{Finetuned on GeoChat} & & & & & & & & & & \\
        LHRS-Bot~\cite{muhtar2024lhrs} & \pz0.2  & \pz1.9 & \pz9.4 & 18.6 & \pz3.8 & \pz5.2 & \pz3.2 & \pz1.2& \pz2.5 & \pz2.7   \\
        MiniGPTv2~\cite{chen2023minigptv2} & \pz1.7 & \pz9.9 & 21.9 & \pz9.1 & \pz3.6 & \pz2.6 & {--}& {--} & \pz8.2 & \pz7.6 \\
        GeoChat~\cite{cvpr2024geochat}    & \pz2.9 & 13.6 & 21.7 & 16.0 & \pz4.3 & 11.8 & {--} & {--} & 10.5 & 10.6 \\
        TeoChat~\cite{irvin2024teochat}  & \pz2.3 & 11.6 & 29.0 & 32.2 & \pz9.3 & 10.6 & 15.1 & \pz4.8 & 11.2 & 11.1 \\
        % Ferret~\cite{you2023ferret}    & \pz7.7 & 25.4 & 32.8 & 23.7 & 16.4 & 17.0 & 27.9 & \pz7.0 & 19.6 & 19.5 \\
        InternVL2-4B~\cite{chen2024internvl} & \pz7.0 & 22.3 & 33.5 & 24.4 & 11.5 & 12.5 & 25.6 & \pz8.4 & 18.9 & 18.2 \\
        EarthDial~\cite{Soni_2025_CVPR} & \pz6.8 & 23.8 & 32.9 & 25.9 & 13.0 & 13.9 & 26.0 & \pz8.4 & 19.3 & 18.6 \\
        % \midrule
        \rowcolor{cyan!15}
        % Ours & \textbf{9.5} & \textbf{32.2} & \textbf{50.9} & \textbf{40.7} & \textbf{13.1} & \textbf{14.3} & \textbf{39.5} & \textbf{9.7} & \textbf{28.0} & \textbf{26.3} \\
        Ours & \textbf{10.1} & \textbf{29.2} & \textbf{44.5} & \textbf{39.0} & \textbf{18.7} & \textbf{20.2} & \textbf{33.8} & \pz\textbf{9.0} & \textbf{24.8} & \textbf{24.4} \\
        % \aboverulesepcolor{cyan!15}
        \midrule
        \venue{Finetuned on EarthDial} & & & & & & & & & & \\
        EarthDial~\cite{Soni_2025_CVPR} & 11.4  & 31.7 & 39.1 & 28.2 & 18.1 & 18.8 & 34.4 & 12.0 & 25.8 & 25.0 \\
        \rowcolor{cyan!15}
        % Ours & \textbf{11.5} & \textbf{36.6} &  \textbf{57.5}& \textbf{32.0} &  \textbf{29.1}& \textbf{29.5} & \textbf{37.5} & 10.9 & \textbf{31.3} & \textbf{31.2} 
        Ours & \textbf{13.5} & \textbf{39.5} &  \textbf{57.9}& \textbf{38.7} &  \textbf{29.7}& \textbf{30.9} & \textbf{38.8} & \textbf{15.9} & \textbf{33.4} & \textbf{33.3}
        \\
        % \aboverulesepcolor{cyan!15}
        \midrule
    \end{tabular}
        }
        \vspace*{-.8\baselineskip}

\end{table}

\noindent{\textbf{Discussion.}} We report quantitative results on the GeoChat test set in~\cref{tab:groundingperformance}. The table details two finetuning experiments. In both cases, our method demonstrates clear performance gains: on the GeoChat dataset, we achieve $24.4\%$ accuracy, outperforming the strongest baseline EarthDial by $5.8\%$. When finetuning on the EarthDial dataset, our model improves the overall accuracy by a sizable margin of $8.3\%$ (relative improvement of $33.2\%$). This robust performance further translates to the zero-shot setting shown in~\cref{tab:referred_object_detection_zeroshot}, where our model outperforms baseline methods on two out of three considered downstream tasks (NWPU VHR-10 and Swimming Pools), while remaining competitive on Tree Crown Detection. We further provide several qualitative examples in~\cref{fig:qualitativegrounding}.

\begin{table*}[t]
\centering
\caption{
    \textbf{Quantitative results for zero-shot referring object detection.} We compare our model's generalization performance against several baselines on three datasets: NWPU VHR-10~\cite{cheng2014multi}, Swimming Pool~\cite{Soni_2025_CVPR}, and Urban Tree Crown Detection~\cite{zamboni2021benchmarking}. 
    %We report the accuracy ($\%$) score, broken down by object size (Small, Medium, Large) and the number of queried objects (Single, Multi). ``Ours'' indicates our model's performance. Best results are in \textbf{bold}.
}
\vspace*{-.4\baselineskip}
\label{tab:referred_object_detection_zeroshot}
\resizebox{\textwidth}{!}{
\begin{tabular}{l ccccc ccccc ccccc}
\hline
 & \multicolumn{5}{c}{NWPU VHR-10~\cite{cheng2014multi}} & \multicolumn{5}{c}{Swimming Pool Dataset~\cite{Soni_2025_CVPR}} & \multicolumn{5}{c}{Urban Tree Crown Detection~\cite{zamboni2021benchmarking}}\\
 & \multicolumn{3}{c}{Object Size} & \multicolumn{2}{c}{Referring} & \multicolumn{3}{c}{Object Size} & \multicolumn{2}{c}{Referring} & \multicolumn{3}{c}{Object Size} & \multicolumn{2}{c}{Referring} \\
\cmidrule(lr){2-4} \cmidrule(lr){5-6} \cmidrule(lr){7-9} \cmidrule(lr){10-11} \cmidrule(lr){12-14} \cmidrule(lr){15-16}
 Model & Small & Medium & Large & Single & Multi & Small & Medium & Large & Single & Multi & Small & Medium & Large & Single & Multi  \\
\hline
GeoChat~\cite{cvpr2024geochat} & \pz2.5 & \pz3.2 & 14.7 & 13.2 & \pz1.9 & \pz0.0 & \pz3.1 & \pz7.3 & \pz1.2 & \pz0.6 &  \pz0.0 & \pz1.8 & \pz8.9 & \pz2.9 & \pz3.1  \\
% TeoChat~\cite{irvin2024teochat} & \pz4.8 & \pz6.2 & \pz8.7 & \pz8.0 & \pz4.7 & \pz2.0 & \textbf{11.4} & 19.2 & \pz8.8 & \pz\textbf{3.1} & - & \pz4.2 & 20.4 & \pz6.5 &  \pz7.1 \\
InternVL2-4B~\cite{chen2024internvl} & \pz7.1 &  12.7 & 25.5 & 23.0 & \pz8.1 & \pz0.6 & \pz6.6 & \pz8.9 & \pz4.5 & \pz0.9 & \pz0.0 & \pz3.2 & 13.4 & \pz5.9 & \pz3.1  \\
InternVL2-8B~\cite{chen2024internvl}  & \pz4.3 & 11.8 & 20.7 & 21.7 & \pz5.9 & \pz0.3 & \pz4.7 & 18.3 & \pz7.6 & \pz0.5 & \pz0.6 & \pz4.0 & 17.1 & \pz7.9 & \pz3.9  \\
% Ferret~\cite{you2023ferret} & 4.9 & 7.7 & 17.4 & 18.4 & 3.1 & 2.8 & 7.5 & 25.8 & \textbf{10.4} & 2.3 & 0.2 & 2.2 & 6.6 & 2.8 & 2.4 \\
EarthDial ~\cite{Soni_2025_CVPR} & 11.7 & 14.2 & 23.1 & 25.4 & \pz8.9  & \pz1.0 & \pz7.4 & 24.9 & \pz8.4 & \pz1.0& \pz\textbf{1.1} & \pz\textbf{7.0} & \textbf{25.7} & \textbf{11.1} & \pz6.7 \\
\rowcolor{cyan!15}
% Ours &  \textbf{14.6} & \textbf{22.9} & \textbf{36.7} & \textbf{29.3} & \textbf{14.9} & \textbf{2.9} & \textbf{8.4} & \textbf{27.2} & \textbf{9.6} & \textbf{3.0}   & 0.3 & 4.9 & 15.4 & 6.1 & 6.1 \\ 
Ours &  \textbf{15.4} & \textbf{23.4} & \textbf{37.9} & \textbf{30.7} & \textbf{14.4} & \pz\textbf{1.8} & \textbf{10.1} & \textbf{33.9} & \textbf{11.0} & \pz\textbf{1.9}   & \pz0.0 & \pz5.8 & 22.9 & \pz8.2 & \pz\textbf{8.0} \\
\bottomrule
\end{tabular}
}
\vspace*{-.8\baselineskip}

\end{table*}

\begin{figure}[!t]
    \centering
    \vspace{1.5em}
    \begin{overpic}[width=0.19\textwidth]{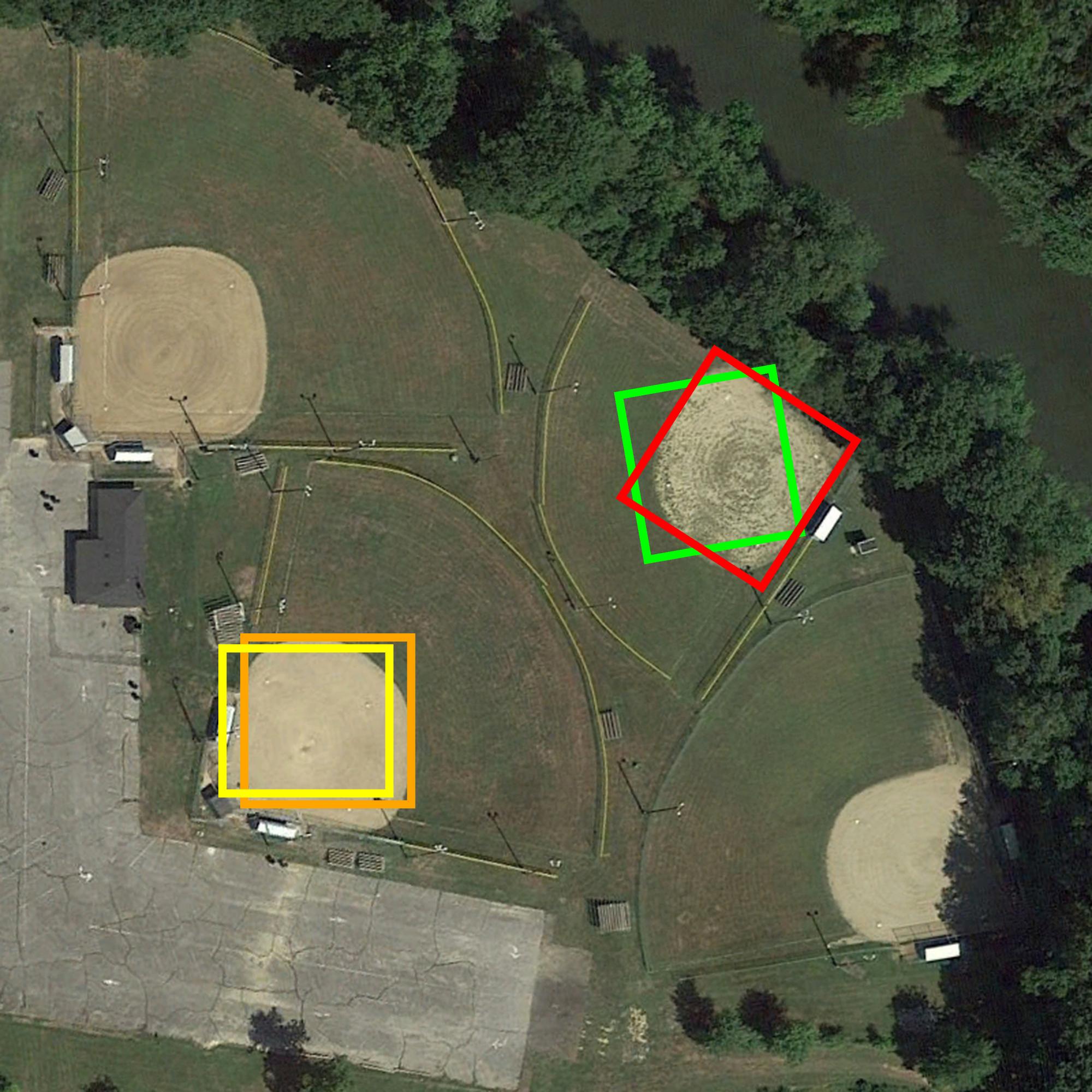}
    \put(0,-12){\begin{minipage}[t]{0.19\textwidth}\tiny\centering ``1 baseballfield at the center"\end{minipage}}
    \end{overpic}
    \begin{overpic}[width=0.19\textwidth]{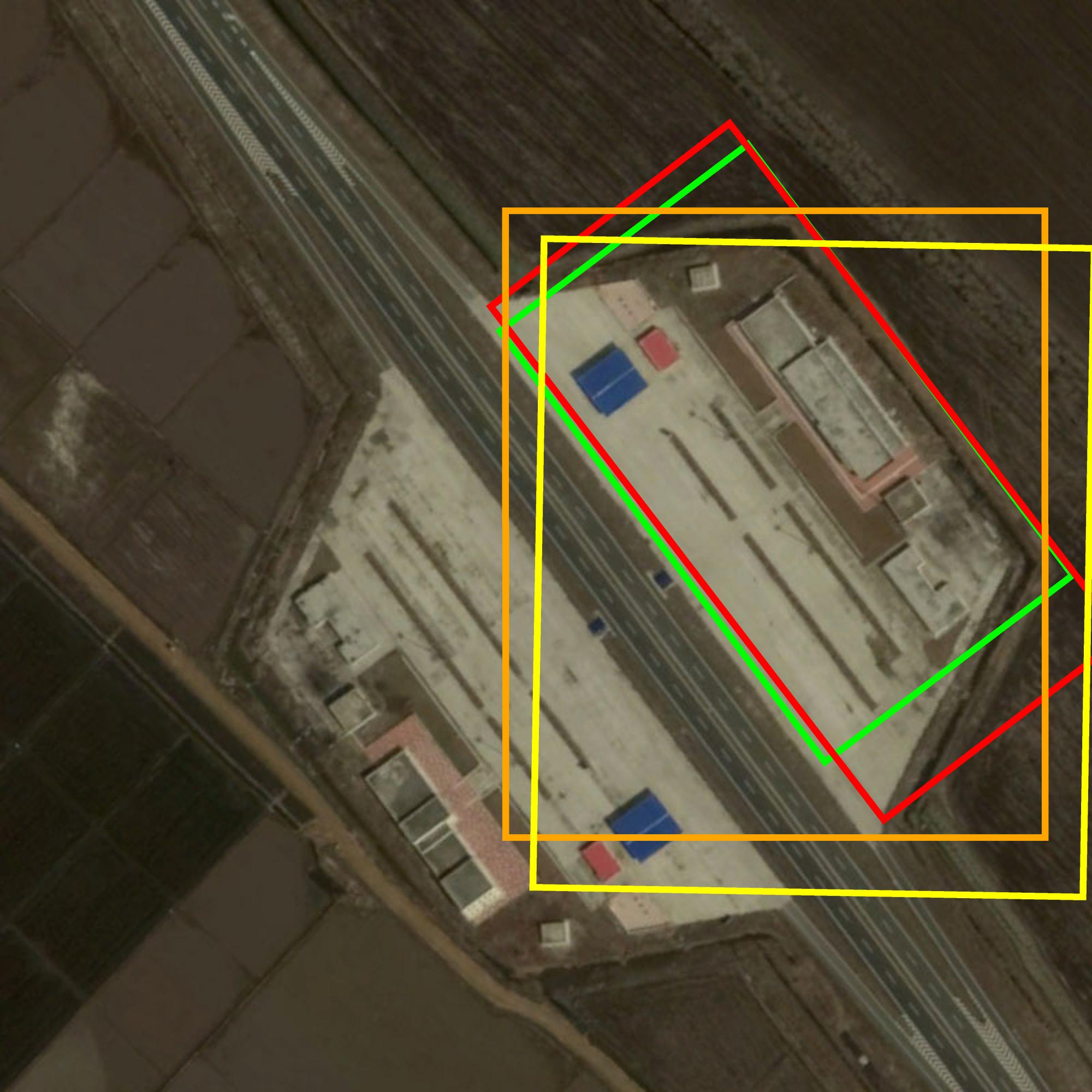}
    \put(0,-12){\begin{minipage}[t]{0.19\textwidth}\tiny\centering ``1 expressway-service-area at the right"\end{minipage}}
    \end{overpic}
    \begin{overpic}[width=0.19\textwidth]{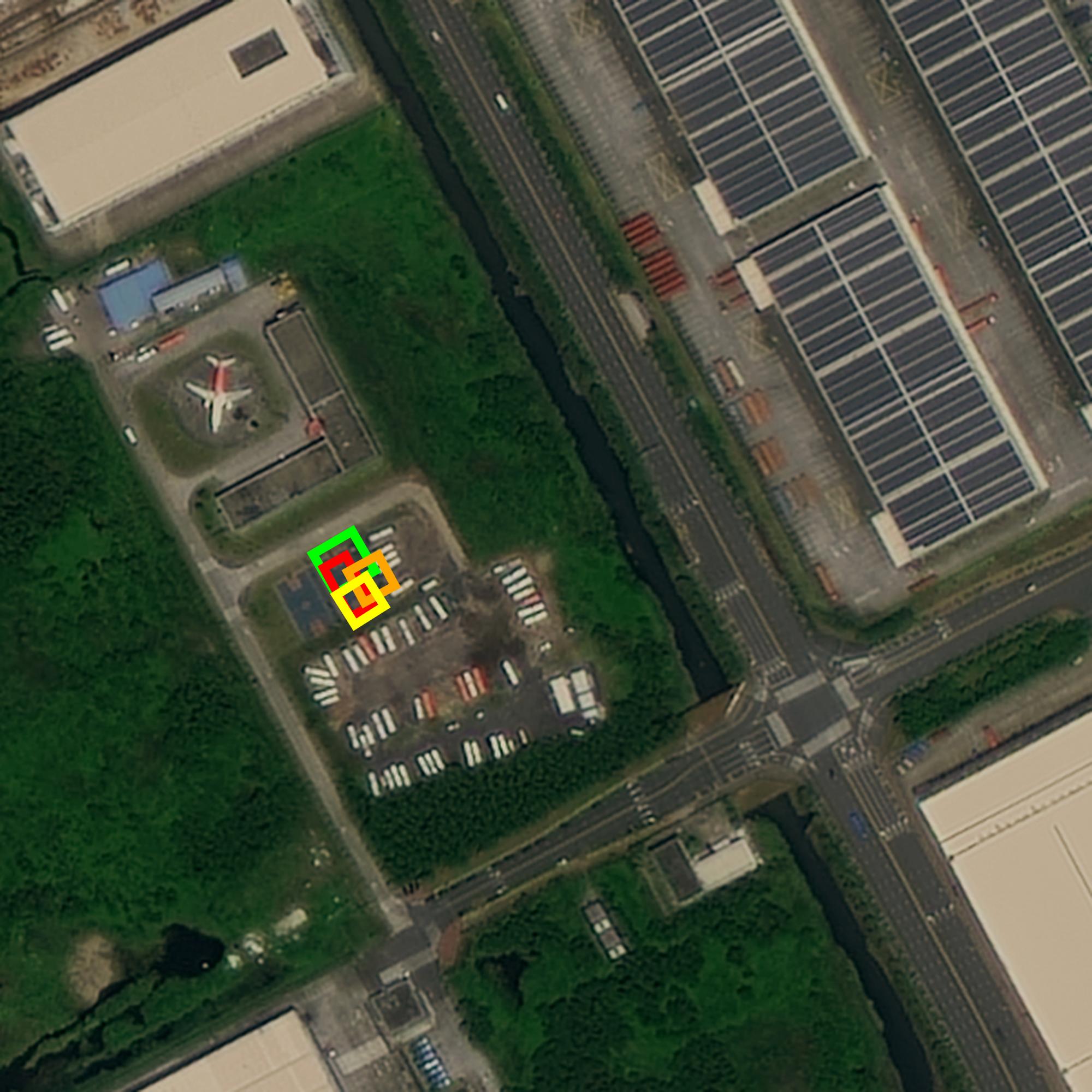}
    \put(0,-12){\begin{minipage}[t]{0.19\textwidth}\tiny\centering ``1 basketball-court at the center"\end{minipage}}
    \end{overpic}
    \begin{overpic}[width=0.19\textwidth]{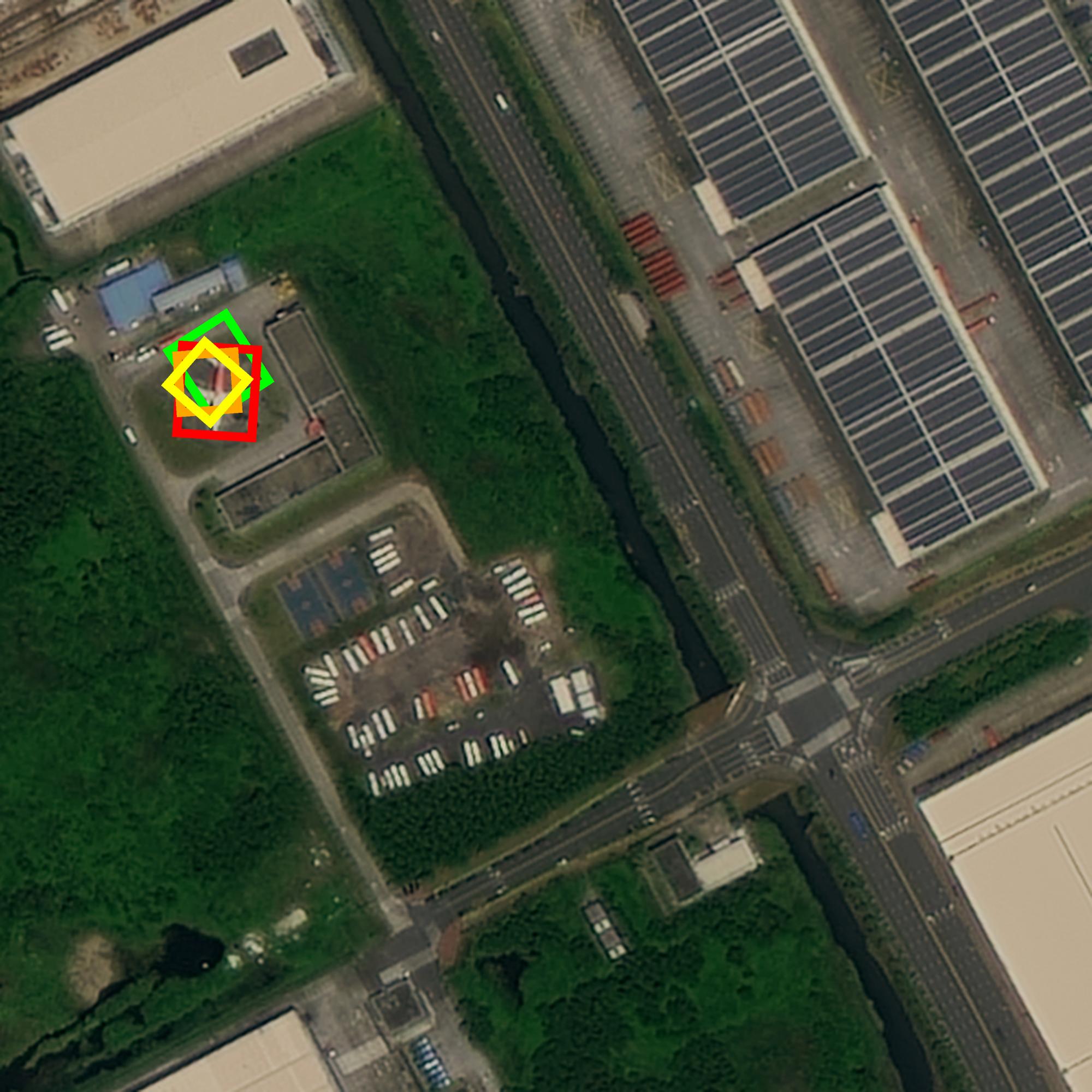}
    \put(0,-12){\begin{minipage}[t]{0.19\textwidth}\tiny\centering ``1 gray airplane at the left of the image"\end{minipage}}
    \end{overpic}
    \begin{overpic}[width=0.19\textwidth]{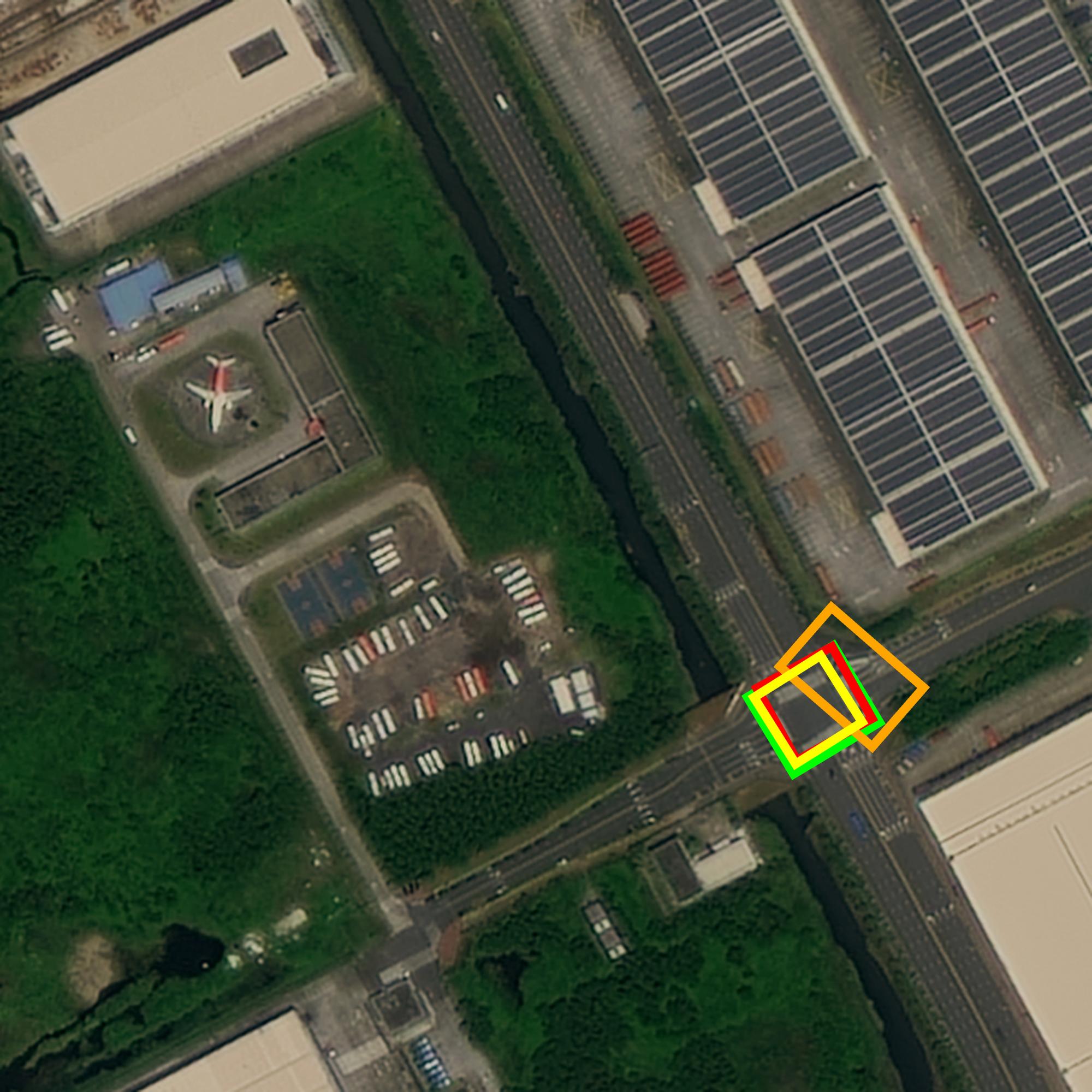}
    \put(0,-12){\begin{minipage}[t]{0.19\textwidth}\tiny\centering ``1 intersection  at the right of the image"\end{minipage}}
    \end{overpic}
     \\[20pt]

    \begin{overpic}[width=0.19\textwidth]{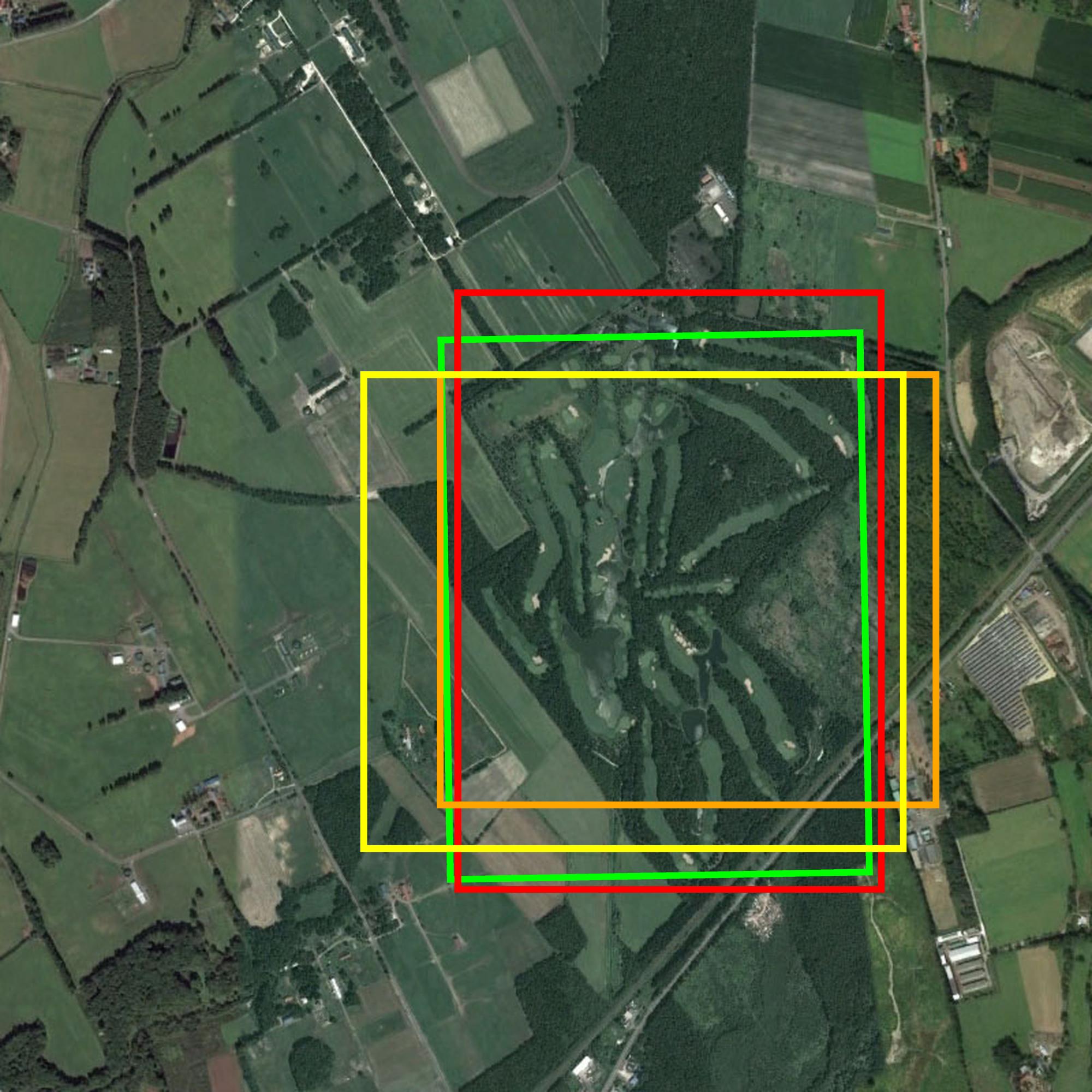}
    \put(0,-12){\begin{minipage}[t]{0.19\textwidth}\tiny\centering ``1 golffield at the center"\end{minipage}}
    \end{overpic}
    \begin{overpic}[width=0.19\textwidth]{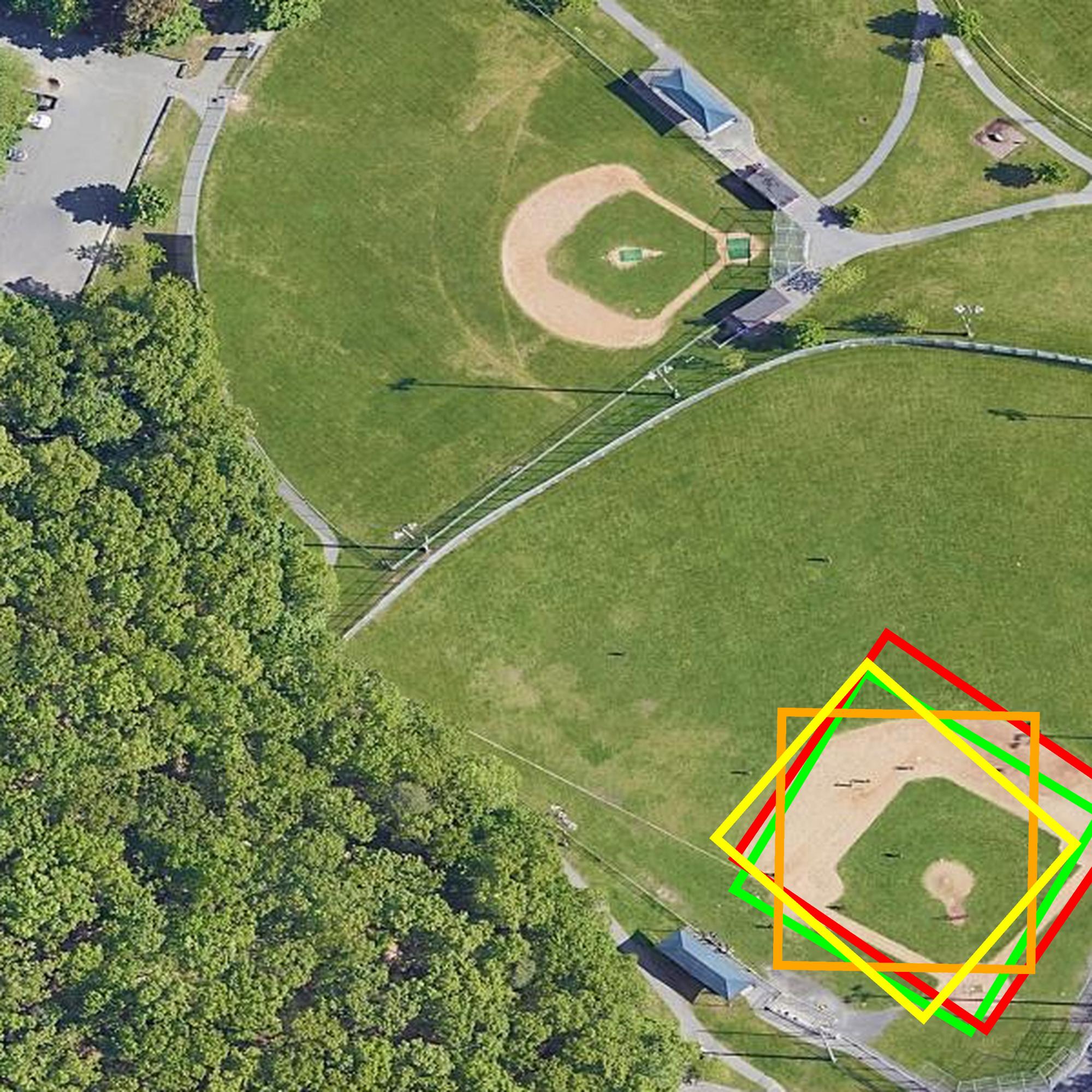}
    \put(0,-12){\begin{minipage}[t]{0.19\textwidth}\tiny\centering ``1 baseball-field at the bottom right"\end{minipage}}
    \end{overpic}
    \begin{overpic}[width=0.19\textwidth]{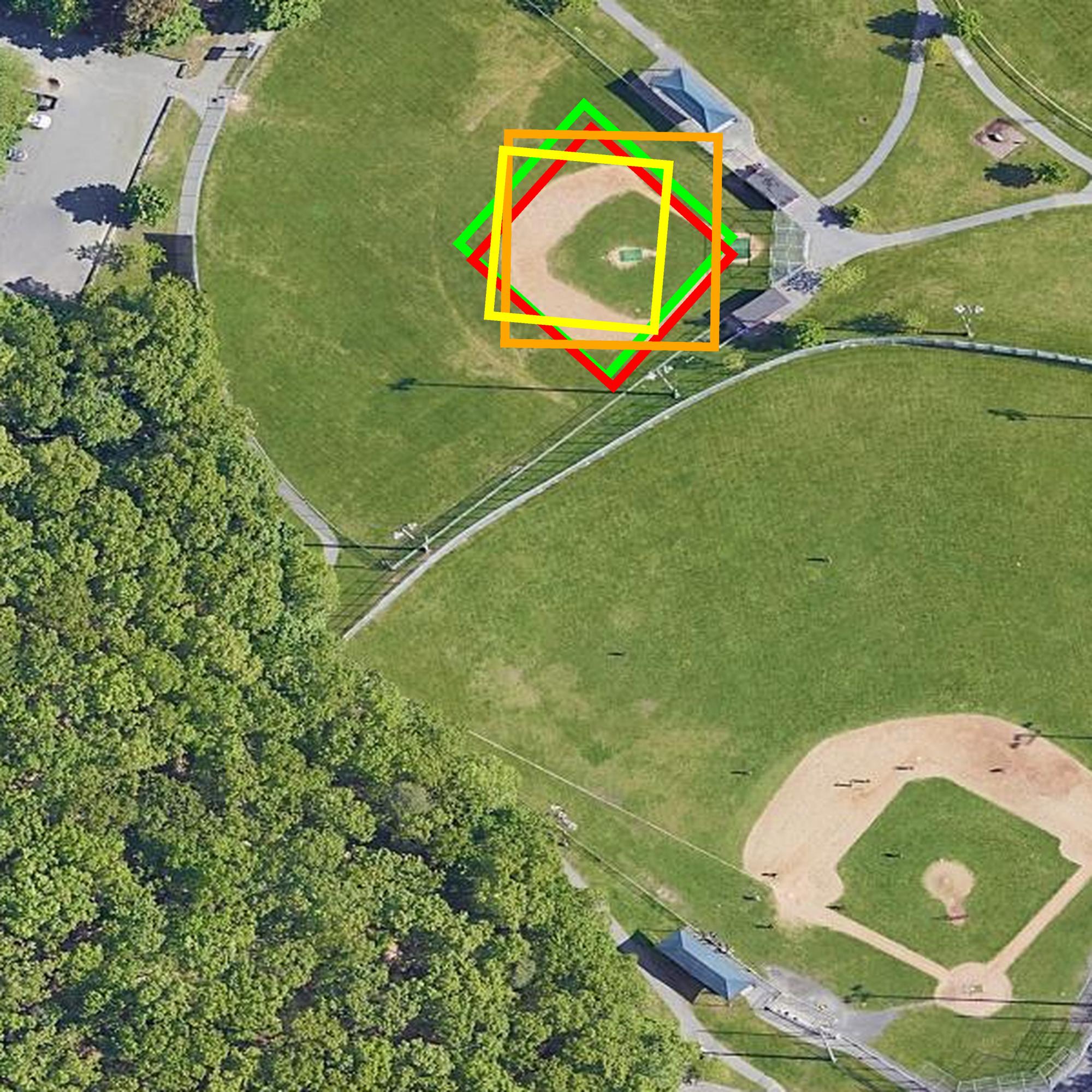}
    \put(0,-12){\begin{minipage}[t]{0.19\textwidth}\tiny\centering ``1 baseball-field at the top"\end{minipage}}
    \end{overpic}
    % \begin{overpic}[width=0.19\textwidth]{figures/internvl_qualitative/sior_1200_visualized.png}
    % \put(0,-8.5){\begin{minipage}[t]{0.19\textwidth}\tiny\centering ``silver airplane"\end{minipage}}
    % \end{overpic}
    \begin{overpic}[width=0.19\textwidth]{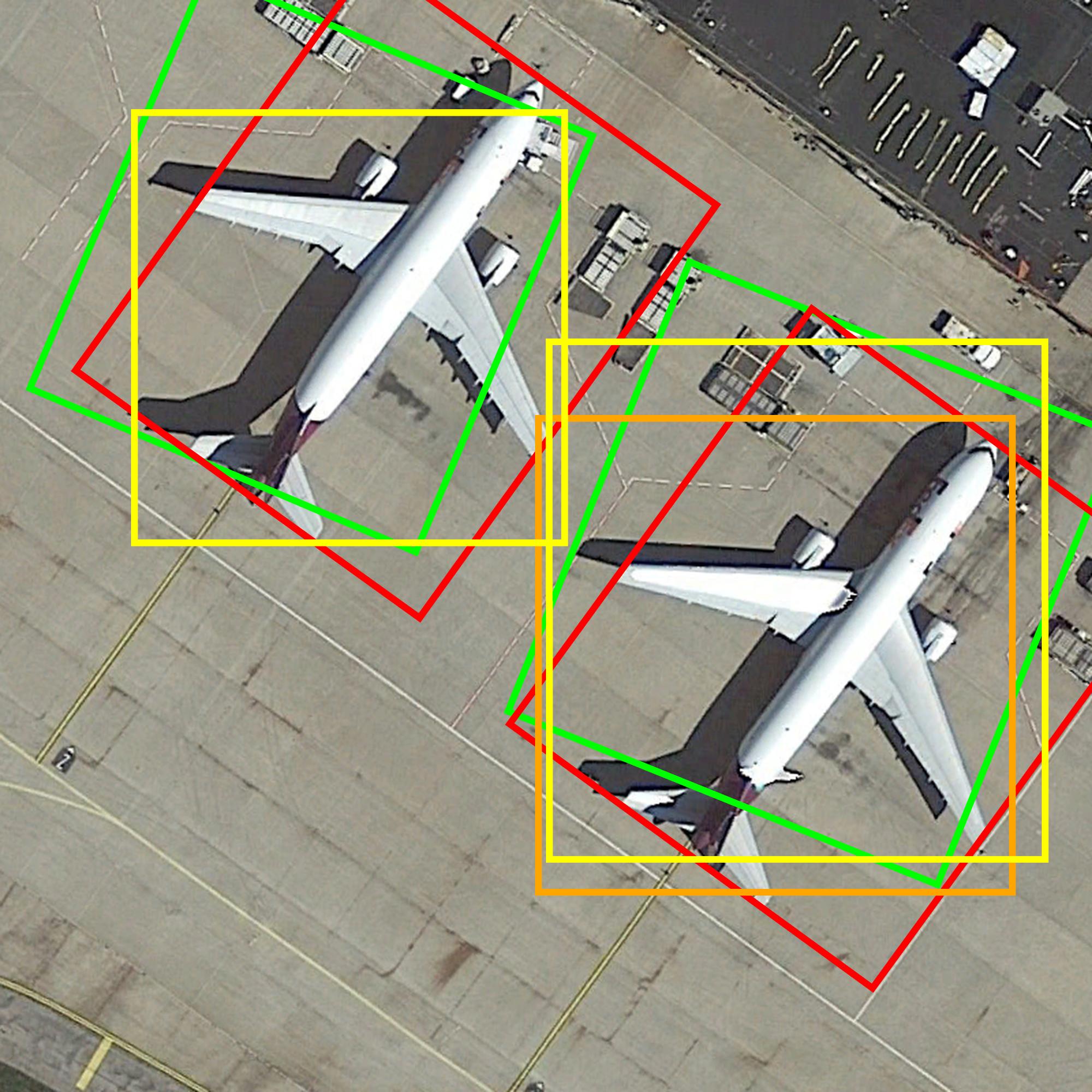}
    \put(0,-12){\begin{minipage}[t]{0.19\textwidth}\tiny\centering ``2 large airplanes"\end{minipage}}
    \end{overpic}
    \begin{overpic}[width=0.19\textwidth]{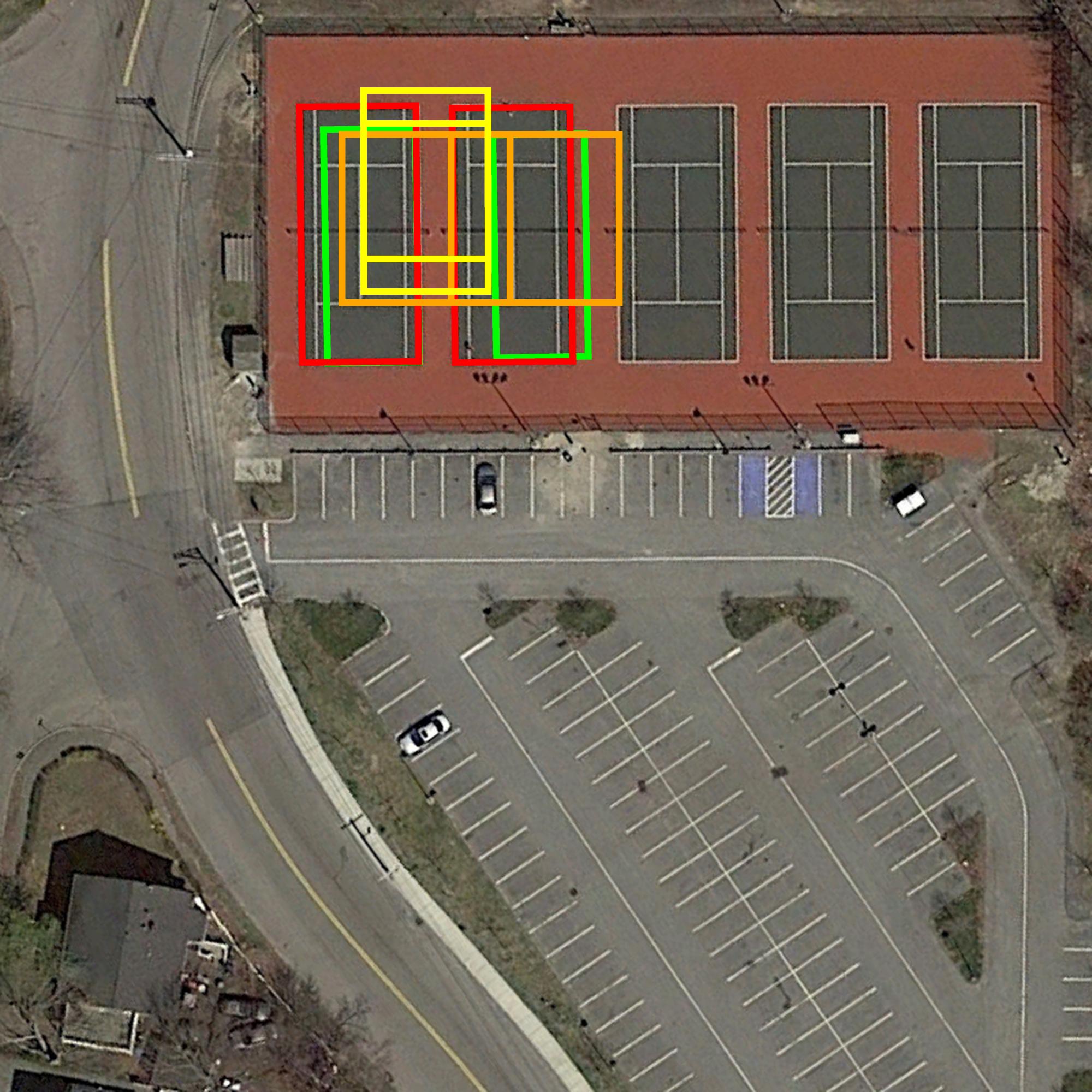}
    \put(0,-12){\begin{minipage}[t]{0.19\textwidth}\tiny\centering ``2 tenniscourts, close to each other at top"\end{minipage}}
    \end{overpic}
     \\[20pt]

    \begin{overpic}[width=0.19\textwidth]{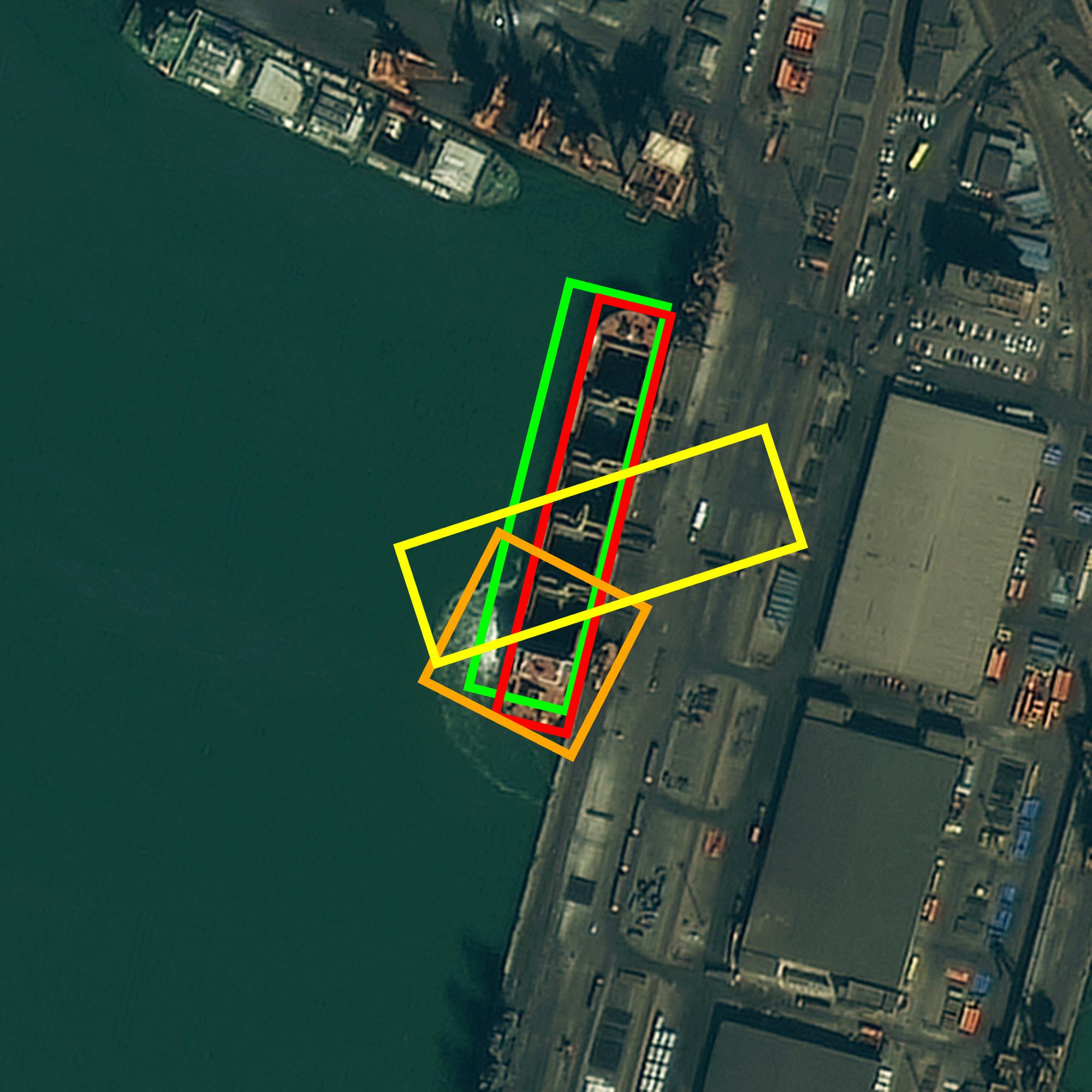}
    \put(0,-12){\begin{minipage}[t]{0.19\textwidth}\tiny\centering ``1 olive large dry-cargo-ship at the center of the image"\end{minipage}}
    \end{overpic}
    \begin{overpic}[width=0.19\textwidth]{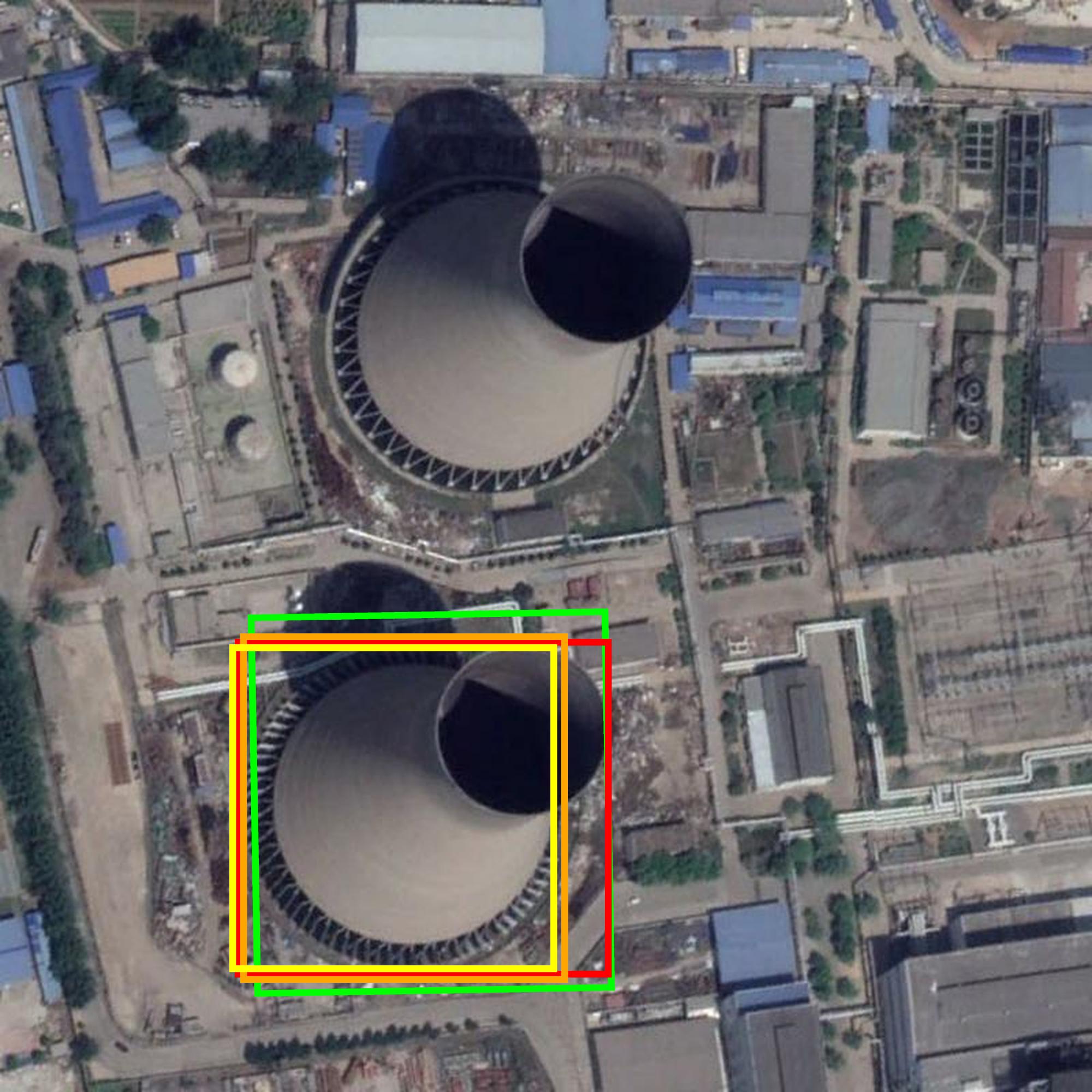}
    \put(0,-12){\begin{minipage}[t]{0.19\textwidth}\tiny\centering ``1 chimney at the bottom"\end{minipage}}
    \end{overpic}
    \begin{overpic}[width=0.19\textwidth]{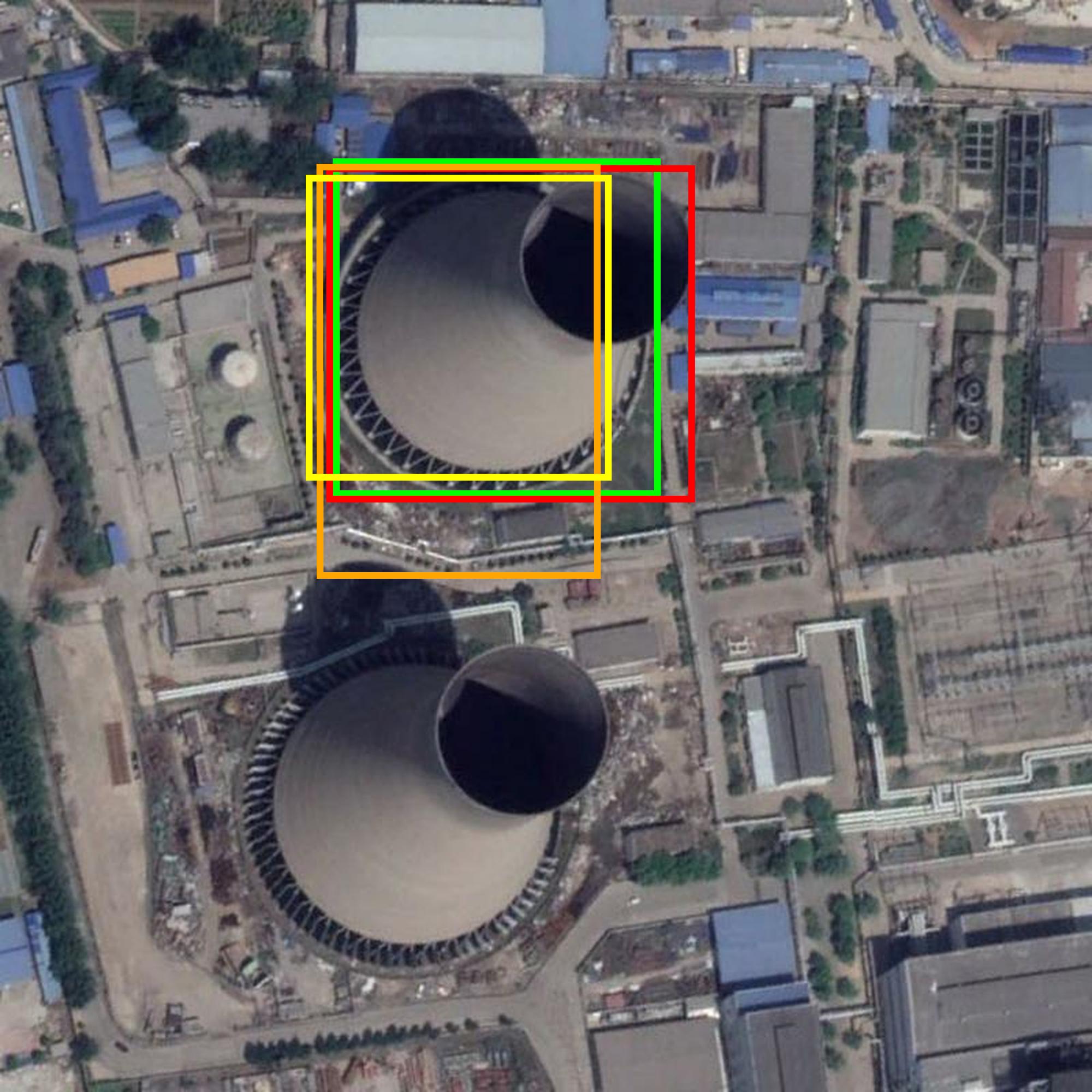}
    \put(0,-12){\begin{minipage}[t]{0.19\textwidth}\tiny\centering ``1 chimney at the center"\end{minipage}}
    \end{overpic}
    \begin{overpic}[width=0.19\textwidth]{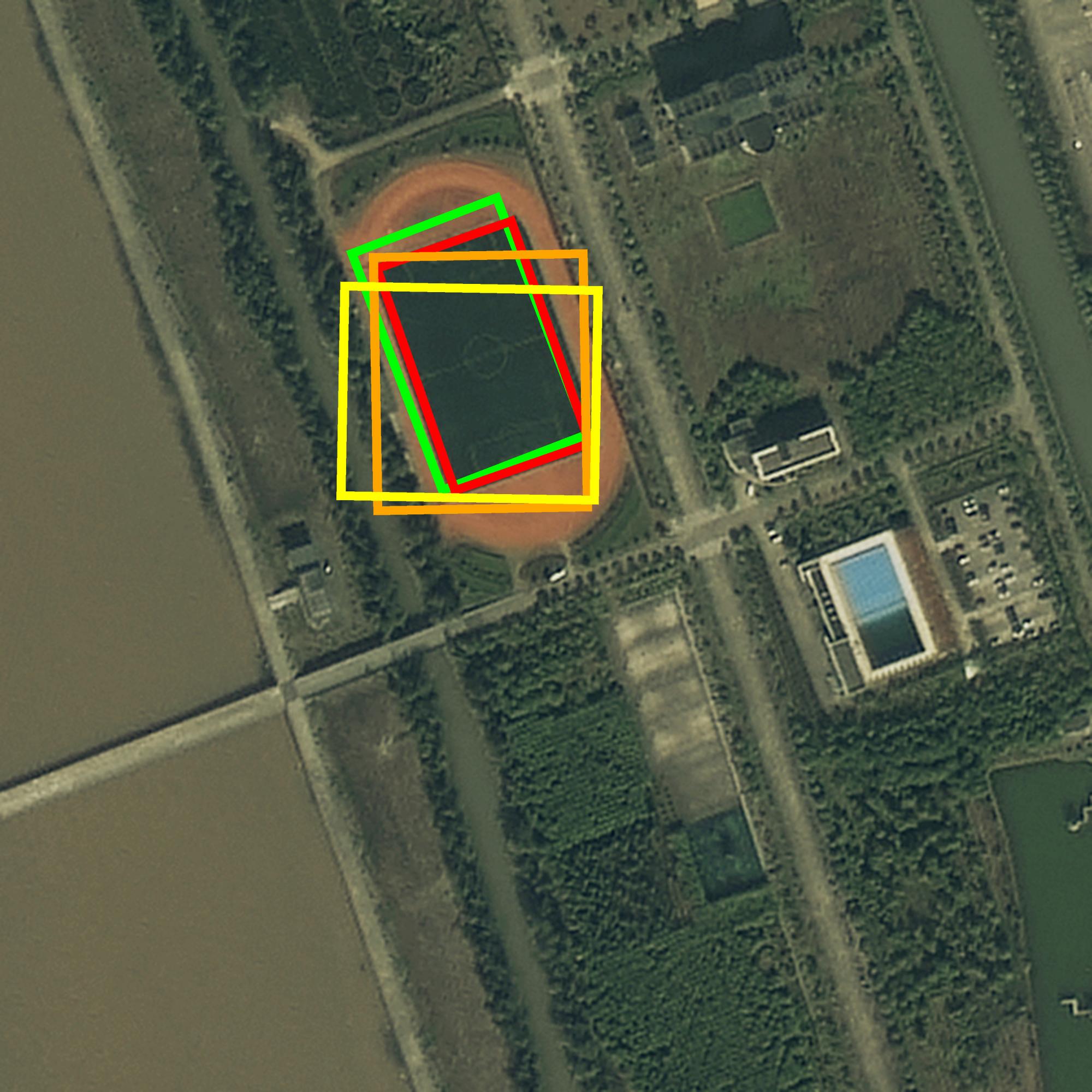}
    \put(0,-12){\begin{minipage}[t]{0.19\textwidth}\tiny\centering ``1 football-field at the center"\end{minipage}}
    \end{overpic}
    \begin{overpic}[width=0.19\textwidth]{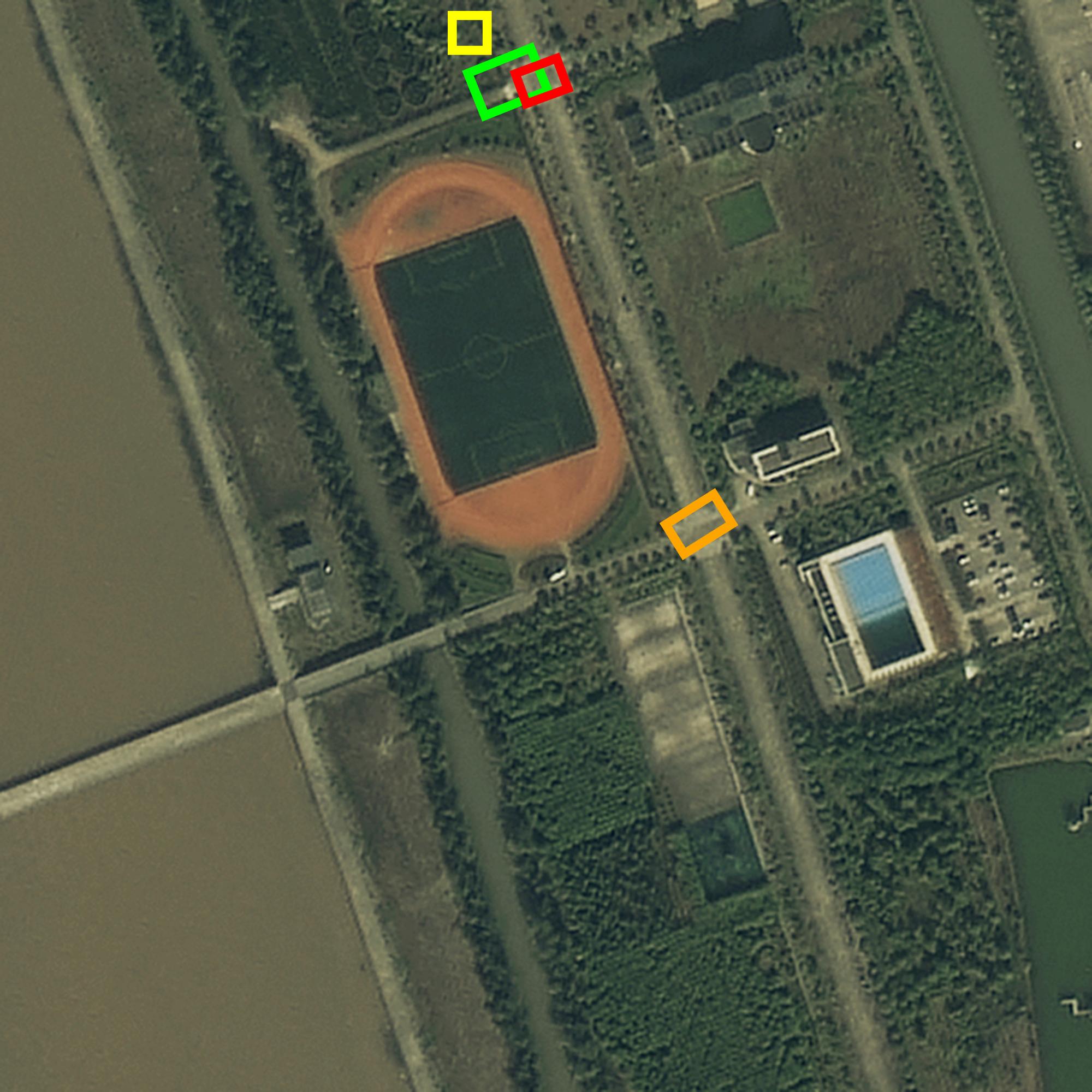}
    \put(0,-12){\begin{minipage}[t]{0.19\textwidth}\tiny\centering ``1 intersection at the top"\end{minipage}}
    \end{overpic}
     \\[20pt]
    \caption{\textbf{Qualitative comparison for visual grounding}. These instances correspond to the quantitative results from~\cref{tab:groundingperformance}. For each sample, we provide predictions by our approach in \textcolor[rgb]{0.0, 1.0, 0.0}{green}, InternVL~\cite{chen2024internvl} in \textcolor[rgb]{1.0, 0.647, 0.0}{orange}, and EarthDial~\cite{Soni_2025_CVPR} in \textcolor[rgb]{1.0, 1.0, 0.0}{yellow}. The ground truth bounding boxes are shown in \textcolor[rgb]{1.0, 0.0, 0.0}{red}.}
    \label{fig:qualitativegrounding}
    \vspace{-1.5em}
\end{figure}

\subsection{Grounding description}

\noindent{\textbf{Datasets.}} We evaluate the grounding description capabilities of our model in a zero-shot setting on four benchmarks: HIT-UAV~\cite{suo2023hit}, NWPU VHR-10~\cite{cheng2014multi}, Swimming Pool~\cite{Soni_2025_CVPR}, and UCAS-AOD~\cite{zhu2015orientation}. These datasets cover diverse object categories and data types ranging from standard imagery to thermal data. 

\noindent{\textbf{Task.}} Grounding description is a challenging task, where the model predicts both a natural language response $\ma$ and corresponding bounding box locations $\mell$. This dual-output structure requires an integrated capacity for both high-level linguistic reasoning and precise localization. We evaluate localization accuracy at IoU thresholds of $0.5$ and $0.25$ ($\text{Acc@}0.5$ and $\text{Acc@}0.25$), and assess the text quality using ROUGE-1 (R-1), ROUGE-L (R-L), and METEOR (MT) scores.

\noindent{\textbf{Discussion.}} We provide quantitative comparisons for zero-shot grounding description in~\cref{tab:grounding_description}. Our model achieves the most consistent performance overall, while outperforming the baseline approaches on ROUGE text metrics (R-1 and R-L) across all four benchmarks. This improvement is particularly notable on UCAS-AOD, where our model achieves a $33.2\%$ R-1 score, a $+12.0\%$ absolute gain over the best baseline EarthDial. Furthermore, our method demonstrates strong localization capabilities, surpassing EarthDial by $+4.2\%$ in $\text{Acc@}0.5$ on the NWPU VHR-10 dataset.
\begin{table}[!t]
    \centering
      \caption{\textbf{Quantitative results for zero-shot grounding description.} We evaluate our model's ability to generate both textual descriptions and corresponding bounding boxes for given text queries and images. Performance is reported with ROUGE-1 (R-1), ROUGE-L (R-L), METEOR (M-T) scores, and accuracy ($\%$) for bounding box prediction (IoU $>0.5$ and $0.25$).}
    \vspace*{-.8\baselineskip}
    \label{tab:grounding_description}
    \resizebox{\textwidth}{!}{
    \begin{tabular}{l ccccc ccccc ccccc ccccc}
        \toprule
          & \multicolumn{5}{c}{HIT-UAV~\cite{suo2023hit}} & \multicolumn{5}{c}{NWPU VHR-10~\cite{cheng2014multi}} & \multicolumn{5}{c}{Swimming Pool Dataset~\cite{Soni_2025_CVPR}} & \multicolumn{5}{c}{UCAS-AOD~\cite{zhu2015orientation}} \\ 
        \cmidrule(lr){2-6} \cmidrule(lr){7-11} \cmidrule(lr){12-16} \cmidrule(lr){17-21}
         Model & \texttt{@}0.5 & \texttt{@}0.25 & R-1 & R-L & MT & \texttt{@}0.5 & \texttt{@}0.25 & R-1 & R-L & MT & \texttt{@}0.5 & \texttt{@}0.25 & R-1 & R-L & MT & \texttt{@}0.5 & \texttt{@}0.25 & R-1 & R-L & MT \\
        \midrule
        
        GPT-4o & \pz0.1 & \pz0.7 & 14.2 & 10.6 & \pz7.2 & \pz0.7 & \pz6.1 & 14.7 & 10.8 & \pz9.4 & \pz0.1 & \pz1.2 & 12.9 & 10.1 & \pz7.8 & \pz0.1 & \pz1.3 & 14.7 & 11.1 & \pz6.0 \\ 
        InternVL2-4B~\cite{chen2024internvl} & \pz0.6 & \pz6.4 & 28.1 & 27.7 & \textbf{24.0} & 10.6 & 29.9 & 30.7 & 29.1 & 21.9 & \pz0.8 & \pz4.2 & 28.3 & 28.1 & 24.6 & \pz4.6 & 31.8 & 21.0 & 20.0 & 11.6\\ 
        GeoChat~\cite{cvpr2024geochat}  & \pz0.8 & \pz8.0 & 22.8 & 22.2 & 22.3 & \pz2.2 & 15.3 & 21.5 & 20.7 & 21.4 & \pz1.8 & \pz\textbf{8.8} & 21.4 & 21.1 & 24.0 & \pz1.4 & 13.6 & 20.0 & 18.8 & 14.2 \\ 
        % TeoChat~\cite{irvin2024teochat}  & \pz1.1 & \pz11.1 & 24.0 & 23.3 & 19.1 & \pz4.5 & 20.9 & 23.6 & 22.6 & 21.2 & \pz1.4 & \pz9.9 & 22.2 & 21.4 & 17.6 & \pz &  &  &  &  \\ 
        % Ferret~\cite{you2023ferret} & \pz1.1 & 13.5 & 31.1 & 30.0 & 19.1 & \pz6.2 & 21.0 & 30.8 & 28.7 & 22.2 & \pz\textbf{2.5} & \textbf{10.8} & 26.1 & 25.3 & 20.9 & \pz2.4 & 20.9 & 33.5 & 30.9 & 13.8\\
        EarthDial ~\cite{Soni_2025_CVPR} & \pz2.6 & 13.9 & 28.3 & 28.1 & 22.2 & 17.1 & \textbf{41.0} & 27.0 & 26.3 & 23.1 & \pz\textbf{1.9} & \pz7.4 & 29.7 & 29.3 & 22.8 & \pz\textbf{8.5} & \textbf{34.0} & 21.2 & 20.3 & 13.0\\
        \rowcolor{cyan!15} \textbf{Ours}  & \pz\textbf{3.4} & \textbf{14.1} & \textbf{30.5} & \textbf{29.8} & 22.5 & \textbf{21.3} & 40.1 & \textbf{35.4} & \textbf{33.5} & \textbf{26.4} & \pz1.2 & \pz7.3 & \textbf{32.1} & \textbf{30.8} & \textbf{25.1} & \pz3.3 & 21.1 & \textbf{33.2} & \textbf{30.9} & \textbf{15.2}\\
        % \aboverulesepcolor{cyan!15}
        %
        \midrule
    \end{tabular}%
    }    
    \vspace*{-.8\baselineskip}

\end{table}

\subsection{Visual question answering}\label{subsec:vqa}
\noindent{\textbf{Datasets.}} We evaluate the performance of our method on visual question answering (VQA) for remote sensing using the HRBEN and LRBEN benchmarks~\cite{lobry2020rsvqa}. Both datasets focus on answering questions about the content of aerial images. On HRBEN, we provide comparisons on the default test set of 62,554 question-answer pairs in a zero-shot manner. We additionally evaluate on the LRBEN test set to provide a comprehensive comparison against existing baselines.

\noindent{\textbf{Tasks.}}
We consider two question types (presence, comparison) for HRBEN and three types (presence, comparison, rural/urban classification) for LRBEN, in alignment with prior work~\cite{hu2023rsgpt,cvpr2024geochat,muhtar2024lhrs}. The goal of presence questions is classifying whether a specific instance is part of an image (e.g., ``Is there a grass area on the right of a commercial building?”). Comparison questions require estimating and contrasting two quantities (e.g., ``Are there more water areas than commercial buildings?” or ``Is the number of buildings equal to the number of roads?”). While presence questions primarily rely on object detection, comparison questions involve both object detection and numerical reasoning, making them inherently more challenging. Rural/urban classification predicts whether an area is urban or rural based on visual criteria such as building density, thus involving scene-level classification rather than object detection.

\begin{table}[t]
\def\arraystretch{1.2}%
\centering
\caption{\textbf{Quantitative VQA results.} The columns report the accuracy for Presence, Comparison, and Rural/Urban question types, along with the average scores.}
\vspace*{-1.2\baselineskip}
% \vspace*{-.4\baselineskip}
\label{tab:vqa_results}
\begin{subtable}[t]{0.35\textwidth}
\centering
\caption{Results on HRBEN.}
\vspace*{-.3\baselineskip}
\label{tab:vqa_hrben}
\resizebox{\columnwidth}{!}{
\begin{tabular}{lccc}
\toprule
 Model & Pres. & Comp. & Avg. \\
\toprule
Qwen-VL~\cite{bai2023qwen} & 66.4 & 60.4 & 63.1\\
LLaVA-1.5~\cite{liu2024improved} & \textbf{69.8} & 67.3 & 68.4 \\
MiniGPTv2~\cite{chen2023minigptv2} & 40.8 & 50.9 & 46.5 \\
GeoChat~\cite{cvpr2024geochat} & 58.5 & 83.2 & 72.3\\
EarthGPT~\cite{zhang2024earthgpt} & 62.8 & 79.5 & 72.1 \\
EarthDial~\cite{Soni_2025_CVPR} & 58.9 & \textbf{83.1} & 72.4 \\
% \rowcolor{cyan!15} Ours & 58.9 & \textbf{83.2} & \textbf{72.5} \\
\rowcolor{cyan!15} Ours & 59.4 & 82.7 & \textbf{72.5} \\
 \bottomrule[0.1em]
\end{tabular}
}

\end{subtable}
\hspace{2em}
\begin{subtable}[t]{0.4\textwidth}
\centering
\caption{Results on LRBEN.}
\vspace*{-.3\baselineskip}
\label{tab:vqa_lrben}
\resizebox{\columnwidth}{!}{
\begin{tabular}{lcccc}
 \toprule
 Model & Pres. & Comp. & R/U & Avg. \\
 \toprule
Qwen-VL~\cite{bai2023qwen} & 38.6 & 67.6 & 61.0 & 55.4\\
LLaVA-1.5~\cite{liu2024improved} & 55.5 & 68.2 & 59.0 & 62.8 \\
MiniGPTv2~\cite{chen2023minigptv2} & 55.2 & 55.5 & 39.0 & 55.0 \\
LHRS-Bot~\cite{muhtar2024lhrs} & 88.5 & 90.0 & 89.1 & 89.4\\
GeoChat~\cite{cvpr2024geochat} & 91.1 & 90.3 & \textbf{94.0} & 90.7\\
EarthDial~\cite{Soni_2025_CVPR} & \textbf{92.6} & \textbf{92.7} & \textbf{94.0} & \textbf{92.7}\\
\rowcolor{cyan!15} Ours & 92.2 & 92.5 & 91.0 & 92.3  \\
 \bottomrule[0.1em]
\end{tabular}

}
\end{subtable}
\vspace*{-.4\baselineskip}
\end{table}

\noindent{\textbf{Discussion.}} In~\cref{tab:vqa_results}, we provide quantitative comparisons for both HRBEN and LRBEN. Our method achieves the best overall zero-shot generalization performance on HRBEN. While LRBEN is fairly saturated and the gap to EarthDial~\cite{Soni_2025_CVPR} is small ($-0.4\%$ average accuracy), our method successfully sets a new state-of-the-art on HRBEN. Ultimately, our approach maintains highly competitive, robust performance across all question types and dataset resolutions. Crucially, this demonstrates that integrating our explicit geometric modeling to enhance spatial localization does not degrade the fundamental reasoning and question-answering capabilities of the vision-language backbone.

\subsection{Ablation study}
In~\cref{subsec:linkingregression}, we introduce our dual visual grounding module, combining next-token prediction and bounding box detection. Here, we assess the impact of key components on the overall performance, see~\cref{tab:ablationstudy} for a summary.

\begin{table}[t]
\def\arraystretch{1.2}%
\centering
\caption{\textbf{Ablation study.} (a) We assess the impact of removing the Hungarian matching, and evaluate how our two-token design compares to merging both tasks into a single token. We further compare our expression-grouped matching defined in~\cref{eq:hungarianloss} to global Hungarian matching ($\checkmark_{glb}$). (b) We quantify the impact of the loss weight $\lambda_\text{bb}$ on visual grounding accuracy. (c) We compare using oriented bounding boxes against dropping the angle at evaluation time, and against training and evaluating with axis-aligned boxes. (d) Moreover, we compare our sigmoid-constrained output with standard $\ell_1$ loss against an unconstrained linear output and a circular $\ell_1$ loss for angle periodicity.}

\vspace*{-.4\baselineskip}
\label{tab:ablationstudy}
\resizebox{\columnwidth}{!}{
\begin{tabular}{ccccccc|cccccc}
\toprule
% Multi-column header grouping the configurations
\multicolumn{2}{c}{(a) Matching} & (b) Weight & \multicolumn{2}{c}{(c) Orientation} & \multicolumn{2}{c}{(d) Grounding} & \multicolumn{6}{c}{} \\
\cmidrule(lr){1-2} \cmidrule(lr){3-3} \cmidrule(lr){4-5} \cmidrule(lr){6-7}
% \cmidrule(l){8-13}
Hung. & $\tokenv{bb}$ & $\lambda_\text{bb}$ & Train w/ & Eval w/ & Loss & w/ Sigmoid & Small & Medium & Large & Grounding & Referring & Overall \\
\midrule

% (a) Matching and Token

           & \checkmark & 10 & \checkmark & \checkmark & $\mathcal{L}_1$ & \checkmark & 11.4  & 37.7   & 54.3    &   28.6  & 31.2   & 31.0  \\ 

$\checkmark$ &            & 10 & \checkmark & \checkmark & $\mathcal{L}_1$ & \checkmark & 10.8    & 35.7    & 53.1    & 26.8    & 29.9    & 29.7    \\ 

$\checkmark_{glb}$      & \checkmark & 10 & \checkmark & \checkmark & $\mathcal{L}_1$ & \checkmark & 12.3 & 37.0 & 55.1 & 28.2 & 31.5 & 31.2  \\ 
           
\midrule

% (b) Lambda Weight
\checkmark & \checkmark & 5  & \checkmark & \checkmark & $\mathcal{L}_1$ & \checkmark & 13.0   & 39.2    & 56.4    & 30.7    & 32.8    & 32.7    \\ 
\checkmark & \checkmark & 20 & \checkmark & \checkmark & $\mathcal{L}_1$ & \checkmark & 12.9   & 40.3  & 55.8    & 30.7    & 33.2    & 33.0    \\ 
\midrule

% (c) Orientation (Checkmarks replaced with explicit 5D/4D values)
\checkmark & \checkmark & 10 & \checkmark &  & $\mathcal{L}_1$ & \checkmark & 11.8 & 33.3 & 40.8 & 24.8 & 26.8 & 26.7 \\ 
\checkmark & \checkmark & 10 &  &    & $\mathcal{L}_1$ & \checkmark & 13.4 & 36.1 & 40.0 & 25.7 & 28.5 & 28.3 \\ 
\midrule

% (d) Loss / Head
\checkmark & \checkmark & 10 & \checkmark & \checkmark & $\mathcal{L}_1$ &      & 11.8 & 36.8 & 55.6 & 30.4 & 31.1 & 31.0 \\ 
\checkmark & \checkmark & 10 & \checkmark & \checkmark & Circ.\ $\mathcal{L}_1$& \checkmark & 12.7 & 35.0 & 39.5 & 24.5 & 27.7 & 27.5 \\ 
\midrule

% Full Baseline (Ours)
\rowcolor{cyan!15} \checkmark & \checkmark & 10 & \checkmark & \checkmark & $\mathcal{L}_1$ & \checkmark & \textbf{13.5} & \textbf{39.5} & \textbf{57.9} & \textbf{30.9} & \textbf{33.4} & \textbf{33.3} \\ 
\bottomrule[0.1em]
\end{tabular}
}
\vspace*{-.8\baselineskip}
\end{table}

\noindent{\textbf{Two token design.}}
We compare our two-token design to a setup where we do not use the $\tokenv{bb}$ token, and instead merge both responsibilities (predicting bounding boxes in the response, producing bounding box coordinates) into a single $\tokenv{loc}$ token. As shown in~\cref{tab:ablationstudy}(a), merging these tasks significantly degrades performance ($29.7\%$ vs. $33.3\%$), since a single embedding vector struggles to represent both the text meaning and the precise geometric coordinates.

\noindent{\textbf{Hungarian matching.}}
We further investigate the role of the Hungarian algorithm in bounding box regression. As shown in~\cref{tab:ablationstudy}(a), applying a global class-agnostic matching scheme ($\checkmark_{glb}$) provides marginal improvement over disabling matching entirely ($31.2\%$ vs. $31.0\%$). In contrast, our expression-grouped matching defined in~\cref{eq:hungarianloss} boosts overall performance to $33.3\%$. This underscores that global assignments struggle to resolve spatial ambiguities, particularly for the grounding task where $95.5\%$ of the samples contain multiple objects.

\noindent{\textbf{Loss weight.}}
We explore the effect of different values for $\lambda_\text{bb}$, which specifies the weight of the grounding loss $\mathcal{L}_\text{ground}$ defined in~\cref{eq:loss}. We find that setting $\lambda_\text{bb}=10$ yields the best overall performance, as shown in~\cref{tab:ablationstudy}(b). 
% Lowering $\lambda_\text{bb}$ to $5$ or increasing it to $20$ slightly reduces the accuracy.

 \noindent{\textbf{Orientation sensitivity.}} To quantify the importance of oriented bounding boxes, we compare our fully oriented predictions against two baselines: (i) training with oriented boxes but ignoring the predicted angle at evaluation time, and (ii) training and evaluating with standard axis-aligned boxes. As shown in~\cref{tab:ablationstudy}(c), the fully oriented approach yields a substantial $+5.0\%$ overall improvement over the axis-aligned variant, with the most pronounced gains on large objects ($+17.9\%$). This confirms that explicit orientation modeling is essential for remote sensing, where objects frequently appear at arbitrary slanted angles.
% such as vehicles, ships, and aircraft 
\noindent{\textbf{Loss function and periodicity.}} In~\cref{tab:ablationstudy}(d), we ablate the grounding head design. Removing the sigmoid constraint for an unconstrained linear head drops performance to $31.0\%$. We also test a circular $\ell_1$ loss to handle $0^\circ/180^\circ$ angular periodicity, formulated as $\mathcal{L}_\text{circ} = \min(|d| \bmod 1, 1-(|d| \bmod 1))$, where $d$ is the normalized angle difference. However, this further decreases performance to $27.5\%$. This suggests that topological boundary cases are sufficiently rare in our dataset, and that the convex stability of the standard $\ell_1$ loss heavily outweighs the theoretical benefits of strict angular periodicity.

\section{Conclusion}
We present \papertitle, a novel approach for spatially-aware visual grounding in remote sensing. Our method integrates multi-modal learning with a specialized localization mechanism, enabling precise alignment of textual queries with visual semantics in satellite imagery. By finetuning a generalist vision-language model with specialized control tokens and a custom grounding module, we bridge the gap between language and spatial reasoning.
Our experiments demonstrate that our model outperforms existing techniques across multiple benchmarks, including referring expression detection, grounding description, and visual question answering. Notably, we achieve a significant $33.2\%$ relative improvement in visual grounding accuracy over prior methods, underscoring the effectiveness of our structured localization approach for real-world satellite data analysis. 
\newpage

\bibliographystyle{splncs04}
\bibliography{main}

\end{document}